%% file: main.tex
\documentclass{article}

\usepackage{microtype}
\usepackage{graphicx}
\usepackage{subfigure}
\usepackage{booktabs} 

\usepackage{hyperref}

\usepackage[accepted]{icml2019}

\icmltitlerunning{Gauge Equivariant CNNs}

\usepackage[utf8]{inputenc}

\usepackage{amsmath}
\usepackage{amsfonts}
\usepackage{amssymb}

\usepackage{wrapfig}
\usepackage{multirow}

\newcommand{\Z}{\mathbb{Z}}
\newcommand{\R}{\mathbb{R}}

\newcommand{\GLR}[1]{\ensuremath{\operatorname{GL}(#1, \mathbb{R})}}

\newcommand{\SO}[1]{\ensuremath{\operatorname{SO}(#1)}}

\newcommand\blfootnote[1]{%
  \begingroup
  \renewcommand\thefootnote{}\footnote{#1}%
  \addtocounter{footnote}{-1}%
  \endgroup
}

\begin{document}

\twocolumn[
\icmltitle{Gauge Equivariant Convolutional Networks and the Icosahedral CNN}



\icmlsetsymbol{equal}{*}

\begin{icmlauthorlist}
\icmlauthor{Taco S. Cohen}{equal,qc}
\icmlauthor{Maurice Weiler}{equal,quva}
\icmlauthor{Berkay Kicanaoglu}{equal,quva}
\icmlauthor{Max Welling}{qc}
\end{icmlauthorlist}

\icmlaffiliation{qc}{Qualcomm AI Research, Amsterdam, NL.}
\icmlaffiliation{quva}{Qualcomm-University of Amsterdam (QUVA) Lab.}

\icmlcorrespondingauthor{Taco S. Cohen}{taco.cohen@gmail.com}
\icmlcorrespondingauthor{Maurice Weiler}{m.weiler@uva.nl}

\icmlkeywords{Geometric Deep Learning, Equivariant Networks, Equivariance, Deep Learning, Machine Learning, Deep Learning, ICML, Gauge Theory, Fiber Bundles}

\vskip 0.3in
]



\printAffiliationsAndNotice{\icmlEqualContribution} 

\begin{abstract}
    The principle of \emph{equivariance to symmetry transformations} enables a theoretically grounded approach to neural network architecture design.
    Equivariant networks have shown excellent performance and data efficiency on vision and medical imaging problems that exhibit symmetries.
    Here we show how this principle can be extended beyond global symmetries to local gauge transformations.
    This enables the development of a very general class of convolutional neural networks on manifolds that depend only on the intrinsic geometry, and which includes many popular methods from equivariant and geometric deep learning.

    We implement gauge equivariant CNNs for signals defined on the surface of the icosahedron, which provides a reasonable approximation of the sphere.
    By choosing to work with this very regular manifold, we are able to implement the gauge equivariant convolution using a single conv2d call, making it a highly scalable and practical alternative to Spherical CNNs.
    Using this method, we demonstrate substantial improvements over previous methods on the task of segmenting omnidirectional images and global climate patterns.
\end{abstract}

\section{Introduction}

By and large, progress in deep learning has been achieved through intuition-guided experimentation.
This approach is indispensable and has led to many successes, but has not produced a deep understanding of \emph{why and when} certain architectures work well.
As a result, every new application requires an extensive architecture search, which comes at a significant labor and energy cost.

Although a theory that tells us which architecture to use for any given problem is clearly out of reach, we can nevertheless come up with \emph{general  principles} to guide architecture search.
One such rational design principle that has met with substantial empirical success  \cite{winkels3DGCNNsPulmonary2018, zaheerDeepSets2017, lunterEquivariantBayesianConvolutional2018}
maintains that network architectures should be equivariant to symmetries.

Besides the ubiquitous translation equivariant CNN, equivariant networks have been developed for sets, graphs, and homogeneous spaces like the sphere (see Sec. \ref{sec:related_work}).
In each case, the network is made equivariant to the global symmetries of the underlying space.
However, manifolds do not in general have global symmetries, and so it is not obvious how one might develop equivariant CNNs for them.

\begin{figure}[t!]
    \centering
    \includegraphics[width=0.46\textwidth]{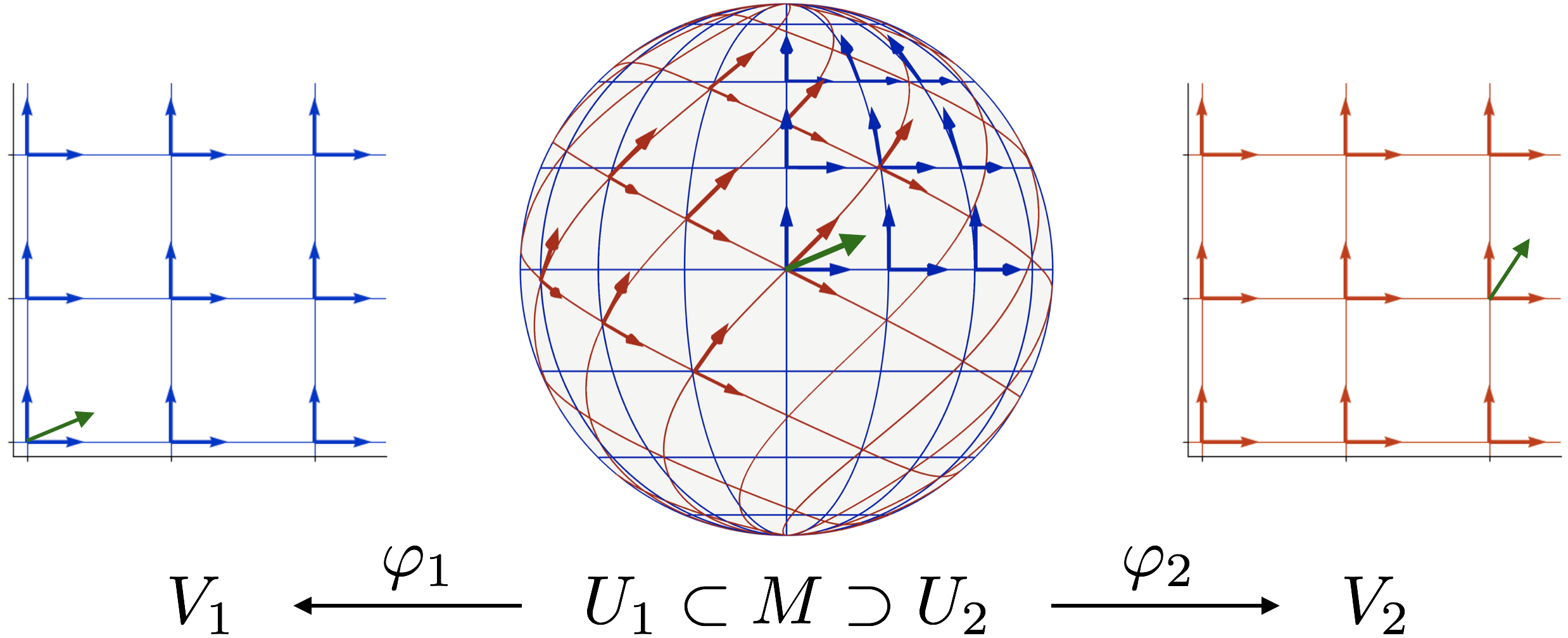}
    \caption{A gauge is a smoothly varying choice of tangent frame on a subset $U$ of a manifold $M$.
    A gauge is needed to represent geometrical quantities such as convolutional filters and feature maps (i.e. fields), but the choice of gauge is ultimately arbitrary.
    Hence, the network should be equivariant to gauge transformations, such as the change between red and blue gauge pictured here.}
    \label{fig:gauge_trafo}
\end{figure}

General manifolds do however have \emph{local gauge symmetries}, and as we will show in this paper, taking these into account is not just useful but \emph{necessary} if one wishes to build manifold CNNs that depend only on the intrinsic geometry.
To this end, we define a convolution-like operation on general manifolds $M$ that is equivariant to local gauge transformations (Fig. \ref{fig:gauge_trafo}).
This \emph{gauge equivariant convolution} takes as input a number of \emph{feature fields} on $M$ of various types (analogous to matter fields in physics), and produces as output new feature fields.
Each field is represented by a number of feature maps, whose activations are interpreted as the coefficients of a geometrical object (e.g. scalar, vector, tensor, etc.) relative to a spatially varying frame (i.e. gauge).
The network is constructed such that if the gauge is changed, the coefficients change in a predictable way so as to preserve their geometrical meaning.
Thus, the search for a geometrically natural definition of ``manifold convolution'', a key problem in geometric deep learning, leads inevitably to gauge equivariance.

Although the theory of gauge equivariant networks developed in this paper is very general, we apply it to one specific manifold: the icosahedron.
This manifold has some global symmetries (discrete rotations), which nicely shows the difference between and interplay of local and global symmetries.
In addition, the regularity and local flatness of this manifold allows for a very efficient implementation using existing deep learning primitives (i.e. conv2d).
The resulting algorithm shows excellent performance and accuracy on segmentation of omnidirectional signals.

Gauge theory plays a central role in modern physics, but has a reputation for being abstract and difficult.
So in order to keep this article accessible to a broad machine learning audience, we have chosen to emphasize geometrical intuition over mathematical formality.

The rest of this paper is organized as follows.
In Sec. \ref{sec:gauge_equivariant_networks},
we motivate the need for working with gauges, and define gauge equivariant convolution for general manifolds and fields.
In section \ref{sec:related_work}, we discuss related work on equivariant and geometrical deep learning.
Then in section \ref{sec:icosahedral_cnns}, we discuss the concrete instantiation and implementation of gauge equivariant CNNs for the icosahedron.
Results on IcoMNIST, climate pattern segmentation, and omnidirectional RGB-D image segmentation are presented in Sec. \ref{sec:experiments}.

\section{Gauge Equivariant Networks}
\label{sec:gauge_equivariant_networks}

Consider the problem of generalizing the classical convolution of two planar signals (e.g. a feature map and a filter) to signals defined on a manifold $M$.
The first and most natural idea comes from thinking of planar convolution in terms of \emph{shifting} a filter over a feature map.
Observing that shifts are symmetries of the plane (mapping the plane onto itself while preserving its structure), 
one is led to the idea of transforming a filter on $M$ by the symmetries of $M$.
For instance, replacing shifts of the plane by rotations of the sphere, one obtains Spherical CNNs \cite{cohenSphericalCNNs2018}.

This approach works for any \emph{homogeneous space}, where by definition it is possible to move from any point $p \in M$ to any other point $q \in M$ using an appropriate symmetry transformation \cite{kondorGeneralizationEquivarianceConvolution2018, cohenIntertwinersInducedRepresentations2018, cohenGeneralTheoryEquivariant2018}.
On less symmetrical manifolds however, it may not be possible to move the filter from any point to any other point by symmetry transformations.
Hence, transforming filters by symmetry transformations will in general not provide a recipe for weight sharing between filters at all points in $M$.

\begin{figure}
  \centering
  \begin{minipage}[c]{0.18\textwidth}
    \includegraphics[width=2.6cm]{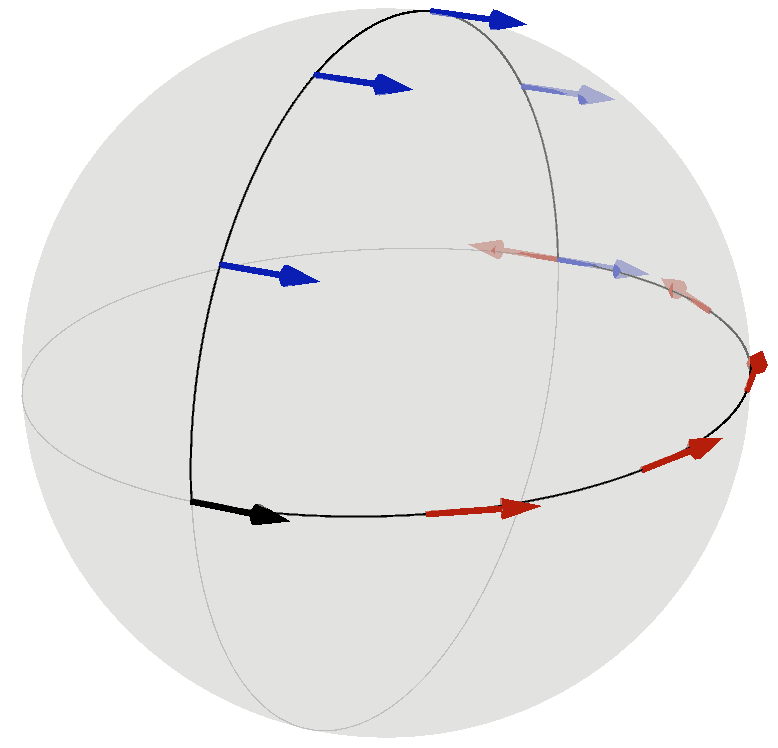}
  \end{minipage}\hfill
  \begin{minipage}[c]{0.3\textwidth}
  \vspace{-0.8 \baselineskip}
  \caption{
       On curved spaces, parallel transport is path dependent. The black vector is transported to the same point via two different curves, yielding different results. The same phenomenon occurs for other geometric objects, including filters.
  }
  \label{fig:parallel_transport}
  \end{minipage}
\end{figure}

Instead of symmetries, one can move the filter by parallel transport \cite{schonsheckParallelTransportConvolution2018}, but as shown in Fig. \ref{fig:parallel_transport}, this leaves an ambiguity in the filter orientation, because parallel transport is \emph{path dependent}.
This can be addressed by using only \emph{rotation invariant} filters \cite{boscainiLearningClassSpecific2015, brunaSpectralNetworksDeep}, albeit at the cost of limiting expressivity.

The key issue is that on a manifold, there is no preferred gauge (tangent frame), relative to which we can position our measurement apparatus (i.e. filters), and relative to which we can describe measurements (i.e. responses).
We must choose a gauge in order to 
numerically represent geometrical quantities and perform 
computations, but since it is arbitrary, the computations should be independent of it.

This does not mean however that the \emph{coefficients} of the feature vectors should be invariant to gauge transformations, but rather that the feature vector itself should be invariant.
That is, a gauge transformation leads to a change of basis $e_i \mapsto \tilde{e}_i$ of the feature space (fiber) at $p \in M$, so the feature vector coefficients $f_i$ should change equivariantly to ensure that the vector $\sum_i f_i e_i = \sum_i \tilde{f}_i \tilde{e}_i$ itself is unchanged.

Before showing how this is achieved, we note that on \emph{non-parallelizable} manifolds such as the sphere, it is not possible to choose a smooth global gauge.
For instance, if we extend the blue gauge pictured in Fig. \ref{fig:gauge_trafo} to the whole sphere, we will innevitably create a singularity where the gauge changes abruptly.
Hence, in order to make the math work smoothly, it is standard practice in gauge theory to work with multiple gauges defined on overlapping charts, as in Fig. \ref{fig:gauge_trafo}.

The basic idea of gauge equivariant convolution is as follows.
Lacking alternative options, we choose arbitrarily a smooth \emph{local} gauge on subsets $U \subset M$ (e.g. the red or blue gauge in Fig. \ref{fig:gauge_trafo}).
We can then position a filter at each point $p \in U$, defining its orientation relative to the gauge.
Then, we match an input feature map against the filter at $p$ to obtain the value of the output feature map at $p$.
For the output to transform equivariantly, certain linear constraints are placed on the convolution kernel.
We will now define this formally.

\subsection{Gauges, Transformations, and Exponential Maps}

We define a gauge as a position-dependent invertible linear map $w_p : \R^d \rightarrow T_p M$, where $T_p M$ is the tangent space of $M$ at $p$.
This determines a frame $w_p(e_1), \ldots, w_p(e_d)$ in $T_p M$, where $\{e_i\}$ is the standard frame of $\R^d$.

A gauge transformation (Fig. \ref{fig:gauge_trafo}) is a position-dependent change of frame, which can be described by maps $g_p \in \GLR{d}$ (the group of invertible $d \times d$ matrices).
As indicated by the subscript, the transformation $g_p$ depends on the position $p \in U \subset M$.
To change the frame, simply compose $w_p$ with $g_p$, i.e. $w_p \mapsto w_p g_p$.
It follows that component vectors $v \in \R^d$ transform as $v \mapsto g_p^{-1} v$, so that the vector $(w_p g_p) (g_p^{-1} v) = w_p v \in T_p M$ itself is invariant.

If we derive our gauge from a coordinate system for $M$ (as shown in Fig. \ref{fig:gauge_trafo}), then a change of coordinates leads to a gauge transformation ($g_p$ being the Jacobian of the coordinate transformation at $p$).
But we can also choose a gauge $w_p$ independent of any coordinate system.

It is often useful to restrict the kinds of frames we consider, for example to only allow right-handed or orthogonal frames.
Such restrictions limit the kinds of gauge transformations we can consider.
For instance, if we allow only right-handed frames, $g_p$ should have positive determinant (i.e. $g_p \in \operatorname{GL}^+(d, \R)$), so that it does not reverse the orientation.
If in addition we allow only orthogonal frames, $g_p$ must be a rotation, i.e. $g_p \in \SO{d}$.

In mathematical terms, $G = \GLR{d}$ is called the \emph{structure group} of the theory, and limiting the kinds of frames we consider corresponds to a \emph{reduction of the structure group} \cite{husemollerFibreBundles1994a}.
Each reduction corresponds to some extra structure that is preserved, such as an \emph{orientation} ($\operatorname{GL}^+(d,\R)$) or \emph{Riemannian metric} ($\SO{d}$).
In the Icosahedral CNN (Fig. \ref{fig:atlas}), we will want to preserve the hexagonal grid structure, which corresponds to a restriction to grid-aligned frames and a reduction of the structure group to $G= C_6$, the group of planar rotations by integer multiples of $2\pi/6$.
For the rest of this section, we will work in the Riemannian setting, i.e. use $G = \SO{d}$.

\begin{figure}
    \centering
    \includegraphics[width=6.5cm]{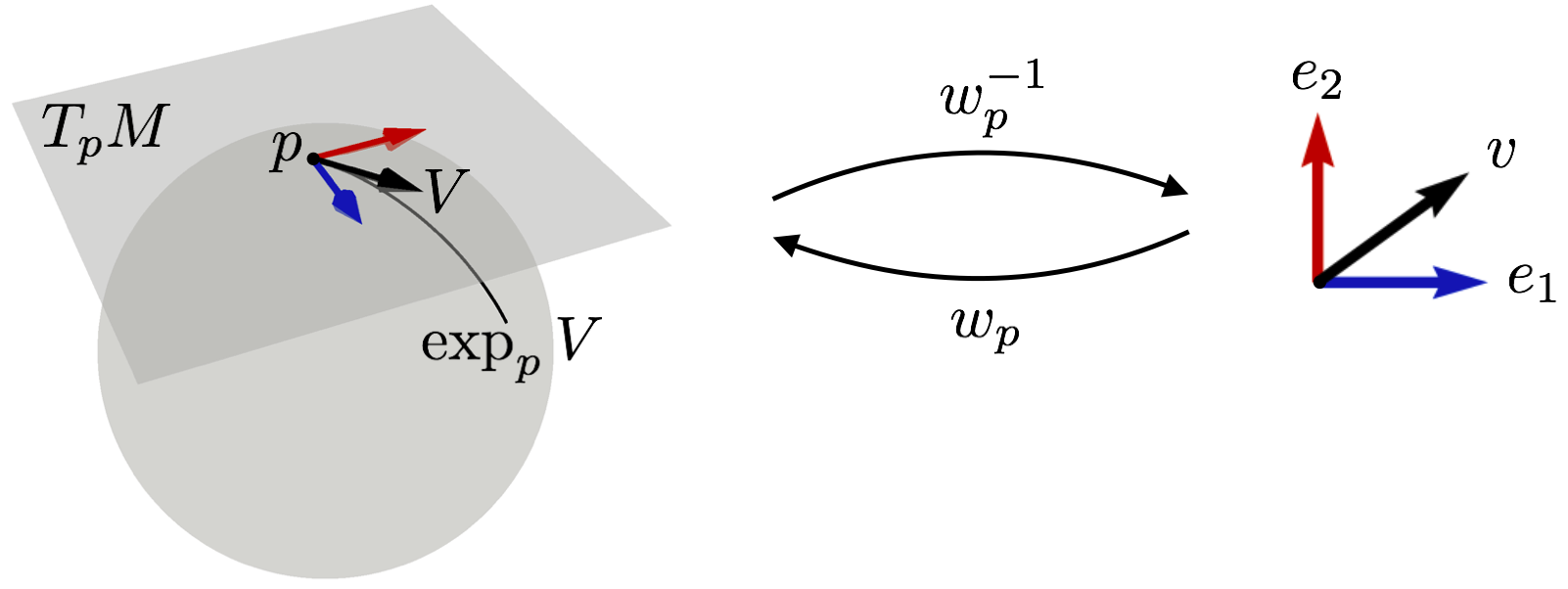}
    \caption{The exponential map and the gauge $w_p$.}
    \label{fig:exponential_map}
\end{figure}

Before we can define gauge equivariant convolution, we will need the exponential map, which gives a convenient parameterization of the local neighbourhood of $p \in M$.
This map $\exp_p : T_p M \rightarrow M$ takes a tangent vector $V \in T_p M$, follows the geodesic (shortest curve) in the direction of $V$ with speed $\|V\|$ for one unit of time, to arrive at a point $q = \exp_p V$ (see Fig. \ref{fig:exponential_map}, \cite{leeIntroductionRiemannianManifolds2018}).

\subsection{Gauge Equivariant Convolution: Scalar Fields}
\label{sec:gconv_local}

Having defined gauges, gauge transformations, and the exponential map, we are now ready to define gauge equivariant convolution.
We begin with scalar input and output fields.

We define a filter as a locally supported function $K: \R^d \rightarrow \R$, where $\R^d$ may be identified with $T_p M$ \emph{via the gauge} $w_p$.
Then, writing $q_v = \exp_p w_p(v)$ for $v \in \R^d$, we define the scalar convolution of $K$ and $f : M \rightarrow \R$ at $p$ as follows:
\begin{equation}
    \label{eq:scalar_conv}
    (K \star f)(p) = \int_{\R^d} K(v) f(q_v) dv.
\end{equation}

The gauge was chosen arbitrarily, so we must consider what happens if we change it.
Since the filter $K : \R^d \rightarrow \R$ is a function of a coordinate vector $v \in \R^d$, and $v$ gets rotated by gauge transformations, the effect of a gauge transformation is a position-dependent rotation of the filters.
For the convolution output to be called a scalar field, it has to be invariant to gauge transformations (i.e. $v \mapsto g_p^{-1} v$ and $w_p \mapsto w_g g_p$).
The only way to make $(K \star f)(p)$ (Eq. \ref{eq:scalar_conv}) invariant to rotations of the filter, is to make the filter rotation-invariant:
\begin{equation}
    \label{eq:scalar_kernel_constraint}
    \begin{aligned}
        \forall g \in G \; : \; K(g^{-1} v) = K(v)  
    \end{aligned}
\end{equation}

Thus, to map a scalar input field to a scalar output field in a gauge equivariant manner, we need to use rotationally symmetric filters.
Some geometric deep learning methods, as well as graph CNNs do indeed use isotropic filters.
However, this is very limiting and as we will now show, unnecessary if one considers non-scalar feature fields.

\subsection{Feature Fields}
\label{sec:feature_fields}

Intuitively, a field is an assignment of some geometrical quantity (feature vector) $f(p)$ of the same type to each point $p \in M$.
The type of a quantity is determined by its transformation behaviour under gauge transformations.
For instance, the word \emph{vector field} is reserved for a field of tangent vectors $v$, that transform like $v(p) \mapsto g_p^{-1} v(p)$ as we saw before.
It is important to note that $f(p)$ is an element of a vector space (``fiber'') $F_p \simeq \R^C$ attached to $p \in M$ (e.g. the tangent space $T_p M$).
All $F_p$ are similar to a canonical feature space $\R^C$, but $f$ can only be considered a function $U \rightarrow \R^C$ \emph{locally}, after we have chosen a gauge, because there is no canonical way to identify all feature spaces $F_p$.

In the general case, the transformation behaviour of a $C$-dimensional geometrical quantity is described by a \emph{representation of the structure group} $G$.
This is a mapping $\rho : G \rightarrow \GLR{C}$ that satisfies $\rho(gh) = \rho(g) \rho(h)$, where $gh$ denotes the composition of transformations in $G$, and $\rho(g) \rho(h)$ denotes multiplication of $C \times C$ matrices $\rho(g)$ and $\rho(h)$.
The simplest examples are the trivial representation $\rho(g) = 1$ which describes the transformation behaviour of scalars, and $\rho(g) = g$, which describes the transformation behaviour of (tangent) vectors.
A field $f$ that transforms like $f(p) \mapsto \rho(g_p^{-1}) f(p)$ will be called a $\rho$-field.

In Section \ref{sec:icosahedral_cnns} on Icosahedral CNNs, we will consider one more type of representation, namely the \emph{regular representation} of $C_6$.
The group $C_6$ can be described as the $6$ planar rotations by $k \cdot 2\pi/6$, or as integers $k$ with addition mod $6$.
Features that transform like the regular representation of $C_6$ are $6$-dimensional, with one component for each rotation.
One can obtain a regular feature by taking a filter at $p$, rotating it by $k \cdot 2 \pi / 6$ for $k=0, \ldots, 5$, and matching each rotated filter against the input signal.
When the gauge is changed, the filter and all rotated copies are rotated, and so the components of a regular $C_6$ feature are cyclically shifted.
Hence, $\rho(g)$ is a $6 \times 6$ cyclic permutation matrix that shifts the coordinates by $k'$ steps for $g = k' \cdot 2 \pi / 6$.
Further examples of representations $\rho$ that are useful in convolutional networks may be found in \cite{cohenSteerableCNNs2017,weiler3DSteerableCNNs2018, thomasTensorFieldNetworks2018, hyPredictingMolecularProperties2018}.

\subsection{Gauge Equivariant Convolution: General Fields}

Now consider a stack of $C_{\text{in}}$ input feature maps on $M$, which represents a $C_\text{in}$-dimensional $\rho_{\text{in}}$-field (e.g. $C_\text{in} = 1$ for a single scalar, $C_\text{in}=d$ for a vector, $C_\text{in} = 6$ for a regular $C_6$ feature, or any multiple of these, etc.).
We will define a convolution operation that takes such a field and produces as output a $C_\text{out}$-dimensional $\rho_\text{out}$-field.
For this we need a filter bank with $C_\text{out}$ output channels and $C_\text{in}$ input channels, which we will describe mathematically as a matrix-valued kernel $K : \R^d \rightarrow \R^{C_\text{out} \times C_\text{in}}$.

We can think of $K(v)$ as a linear map from the input feature space (``fiber'') at $p$ to the output feature space at $p$, these spaces being identified with $\R^{C_\text{in}}$ resp. $\R^{C_\text{out}}$ by the choice of gauge $w_p$ at $p$.
This suggests that we need to modify Eq. \ref{eq:scalar_conv} to make sure that the kernel matrix $K(v)$ is multiplied by a feature vector at $p$, not one at $q_v = \exp_p w_p(v)$.
This is achieved by transporting $f(q_v)$ to $p$ along the unique\footnote{For points that are close enough, there is always a unique geodesic. Since the kernel has local support, $p$ and $q_v$ will be close for all non-zero terms.} geodesic connecting them, before multiplying by $K(v)$.

As $f(q_v)$ is transported to $p$, it undergoes a transformation which will be denoted $g_{p \leftarrow q_v} \in G$ (see Fig. \ref{fig:parallel_transport}).
This transformation acts on the feature vector $f(q_v) \in \R^{C_\text{in}}$ via the representation $\rho_\text{in}(g_{p \leftarrow q_v}) \in \R^{C_\text{in} \times C_\text{in}}$.
Thus, we obtain the generalized form of Eq. \ref{eq:scalar_conv} for general fields:
\begin{equation}
    \label{eq:parallel_transport_conv}
    (K \star f)(p) = \int_{\R^d} K(v) \rho_\text{in}(g_{p \leftarrow q_v} ) f(q_v) dv.
\end{equation}

Under a gauge transformation, we have:
\begin{equation}
    \label{eq:subst}
    \begin{aligned}
        v &\mapsto g_p^{-1} v, \;\;\;\;\; & f(q_v) &\mapsto \rho_\text{in}(g_{q_v}^{-1}) f(q_v), \\
        w_p & \mapsto w_p g_p, \;\;\;\;\; & g_{p \leftarrow q_v} &\mapsto g_p^{-1} g_{p \leftarrow q_v} g_{q_v}.
    \end{aligned}
\end{equation}
For $K\star f$ to be well defined as a $\rho_\text{out}$-field, we want it to transform like $(K \star f)(p) \mapsto \rho_\text{out}(g_p^{-1}) (K \star f)(p)$.
Or, in other words, $\star$ should be gauge equivariant.
This will be the case if and only if $K$ satisfies
\begin{equation}
    \label{eq:kernel_constraint}
    \forall g \in G \,:\,  K(g^{-1} v) = \rho_\text{out}(g^{-1}) K(v) \rho_\text{in}(g).
\end{equation}
One may verify this by making the substitutions of Eq. \ref{eq:subst} in Eq. \ref{eq:parallel_transport_conv} and simplifying using $\rho(gh) = \rho(g) \rho(h)$ and Eq. \ref{eq:kernel_constraint}, to find that $(K \star f)(p) \mapsto \rho_\text{out}(g_p^{-1}) (K\star f)(p)$.

We note that equations \ref{eq:scalar_conv} and \ref{eq:scalar_kernel_constraint} are special cases of \ref{eq:parallel_transport_conv} and \ref{eq:kernel_constraint} for $\rho_\text{in}(g) = \rho_\text{out}(g) = 1$, i.e. for scalar fields.

This concludes our presentation of the general case.
A gauge equivariant $\rho_1 \rightarrow \rho_2$ convolution on $M$ is defined relative to a local gauge by Eq. \ref{eq:parallel_transport_conv}, where the kernel satisfies the equivariance constraint of Eq. \ref{eq:kernel_constraint}.
By defining gauges on local charts $U_i \subset M$ that cover $M$ and convolving inside each one, we automatically get a globally well-defined operation, because switching charts corresponds to a gauge transformation (Fig. \ref{fig:gauge_trafo}), and the convolution is gauge equivariant.

\subsection{Locally Flat Spaces}
\label{sec:locally_flat_spaces}

On flat regions of the manifold, the exponential parameterization can be simplified to $\varphi(\exp_p w_p(v)) = \varphi(p) + v$ if we use an appropriate local coordinate $\varphi(p) \in \R^d$ of $p \in M$.
Moreover, in such a flat chart, parallel transport is trivial, i.e. $g_{p \leftarrow q_v}$ equals the identity.
Thus, on a flat region, our convolution boils down to a standard convolution / correlation:
\begin{equation}
    (K \star f)(x) = \int_{\R^d} K(v) f(x + v) dv.
\end{equation}
Moreover, we can recover group convolutions, spherical convolutions, and convolution on other homogeneous spaces as special cases as well (see supplementary material).

\section{Related work}
\label{sec:related_work}

\paragraph{Equivariant Deep Learning}
Equivariant networks have been proposed for permutation-equivariant analysis and prediction of sets 
\cite{zaheerDeepSets2017, hartfordDeepModelsInteractions2018}, graphs \cite{kondorCovariantCompositionalNetworks2018, hyPredictingMolecularProperties2018, maronInvariantEquivariantGraph2019}, translations and rotations of the plane and 3D space \cite{oyallonDeepRotoTranslationScattering2015, cohenGroupEquivariantConvolutional2016, cohenSteerableCNNs2017, marcosRotationEquivariantVector2017, weilerLearningSteerableFilters2018, weiler3DSteerableCNNs2018, worrallHarmonicNetworksDeep2017, worrallCubeNetEquivariance3D2018, winkels3DGCNNsPulmonary2018, veelingRotationEquivariantCNNs2018, thomasTensorFieldNetworks2018, bekkersRotoTranslationCovariantConvolutional2018, hoogeboomHexaConv2018}, and the sphere (see below).
\citet{ravanbakhshEquivarianceParameterSharing2017} studied finite group equivariance.
Equivariant CNNs on homogeneous spaces were studied by \cite{kondorGeneralizationEquivarianceConvolution2018} (scalar fields) and \cite{cohenIntertwinersInducedRepresentations2018, cohenGeneralTheoryEquivariant2018} (general fields).
In this paper we generalize G-CNNs from homogeneous spaces to general manifolds.

\paragraph{Geometric Deep Learning}

Geometric deep learning \cite{bronsteinGeometricDeepLearning2016} is concerned with the generalization of (convolutional) neural networks to manifolds.
Many definitions of manifold convolution have been proposed, and some of them (those called ``intrinsic'') are gauge equivariant (although to the best of our knowledge, the relevance of gauge theory has not been observed before).
However, these methods are all limited to particular feature types $\rho$ (typically scalar), and/or use a parameterization of the kernel that is not maximally flexible.

\citet{brunaSpectralNetworksDeep, boscainiLearningClassSpecific2015} propose to use isotropic (spectral) filters (i.e. scalar field features), while \cite{masciGeodesicConvolutionalNeural2015} define a convolution that is essentially the same as our scalar-to-regular convolution, followed by a max-pooling over orientations, which in our terminology maps a regular field to a scalar field.
As shown experimentally in \cite{cohenGroupEquivariantConvolutional2016, cohenSteerableCNNs2017} and in this paper, it is often more effective to use convolutions that preserve orientation information (e.g. regular to regular convolution).
Another solution is to align the filter with the maximum curvature direction \cite{boscainiLearningShapeCorrespondence2016}, but this approach is not intrinsic and does not work for flat surfaces or uniformly curved spaces like spheres.

\cite{poulenardMultidirectionalGeodesicNeural2018a} define a multi-directional convolution for ``directional functions'' (somewhat similar to what we call regular fields), but they parameterize the kernel by a \emph{scalar} function on the tangent space, which is very limited compared to our matrix-valued kernel (which is the most general kernel mapping $\rho_1$ fields to $\rho_2$ fields).

\paragraph{Spherical CNNs}
Besides the general theoretical framework of gauge equivariant convolution, we present in this paper a specific model (the Icosahedral CNN), which can be viewed as a fast and simple alternative to Spherical CNNs \cite{cohenSphericalCNNs2018, estevesLearningEquivariantRepresentations2018, boomsmaSphericalConvolutionsTheir2017, suLearningSphericalConvolution2017, perraudinDeepSphereEfficientSpherical2018, jiangSphericalCNNsUnstructured2018, kondorClebschGordanNets2018}.
\citet{liuDeepLearning3D2019} use a spherical grid based on a subdivision of the icosahedron, and convolve over it using a method that is similar to the one presented in Sec. \ref{sec:icosahedral_cnns} (and thus ignores curvature), but this method is not equivariant and does not take into account gauge transformations.
We show in Sec. \ref{sec:experiments} that both are important for optimal performance.

\paragraph{Mathematics \& physics}
To deeply understand gauge equivariant networks, we recommend studying the mathematics of gauge theory: principal \& associated fiber bundles \cite{schullerLecturesGeometricalAnatomy2016, husemollerFibreBundles1994a, steenrodTopologyFibreBundles}.
The work presented in this paper can be understood as replacing the principal $G$-bundle $H \rightarrow H/G$ used in G-CNNs over homogeneous spaces $H/G$ \cite{cohenGeneralTheoryEquivariant2018} by the frame bundle of $M$, which is another principal $G$-bundle.
More details can be found in the supplementary material.

\section{Icosahedral CNNs}
\label{sec:icosahedral_cnns}

In this section we will describe a concrete method for performing gauge equivariant convolution on the icosahedron.
The very special shape of this manifold makes it possible to implement gauge equivariant convolution in a way that is both numerically convenient (no interpolation is required), and computationally efficient (the heavy lifting is done by a single conv2d call).

\subsection{The Icosahedron}
\label{sec:the_icosahedron}

The icosahedron is a regular solid with $20$ faces, $30$ edges, and $12$ vertices (see Fig. \ref{fig:atlas}, left).
It has $60$ rotational symmetries.
This symmetry group will be denoted\footnote{As an abstract group, $\mathcal{I} \simeq A5$ (the alternating group A5), but we use $\mathcal{I}$ to emphasize that it is realized by a set of 3D rotations.} $\mathcal{I}$.

\subsection{The Hexagonal Grid}
\label{sec:grid}

Whereas general manifolds, and even spheres, do not admit completely regular and symmetrical pixelations, we can define an almost perfectly regular grid of pixels on the icosahedron.
This grid is constructed through a sequence of grid-refinement steps.
We begin with a grid $\mathcal{H}_0$ consisting of the corners of the icosahedron itself.
Then, for each triangular face, we subdivide it into 4 smaller triangles, thus introducing 3 new points on the center of the edges of the original triangle.
This process is repeated $r$ times to obtain a grid $\mathcal{H}_r$ with $N = 5 \times 2^{2r+1} + 2$ points (Fig. \ref{fig:atlas}, left).

Each grid point (pixel) in the grid has 6 neighbours, except for the corners of the icosahedron, which have $5$.
Thus, one can think of the non-corner grid points as hexagonal pixels, and the corner points as pentagonal pixels.

Notice that the grid $\mathcal{H}_r$ is perfectly symmetrical, which means that if we apply an icosahedral symmetry $g \in \mathcal{I}$ to a point $p \in \mathcal{H}_r$, we will always land on another grid point, i.e. $gp \in \mathcal{H}_r$.
Thus, in addition to talking about gauge equivariance, for this manifold / grid, we can also talk about (exact) equivariance to \emph{global transformations} (3D rotations in $\mathcal{I}$).
Because these global symmetries act by permuting the pixels and changing the gauge, one can see that a gauge equivariant network is automatically equivariant to global transformations.
This will be demonstrated in Section \ref{sec:experiments}.

\subsection{The Atlas of Charts}
\label{sec:atlas}

\begin{figure}[h!]
    \centering
    \includegraphics[width=8cm]{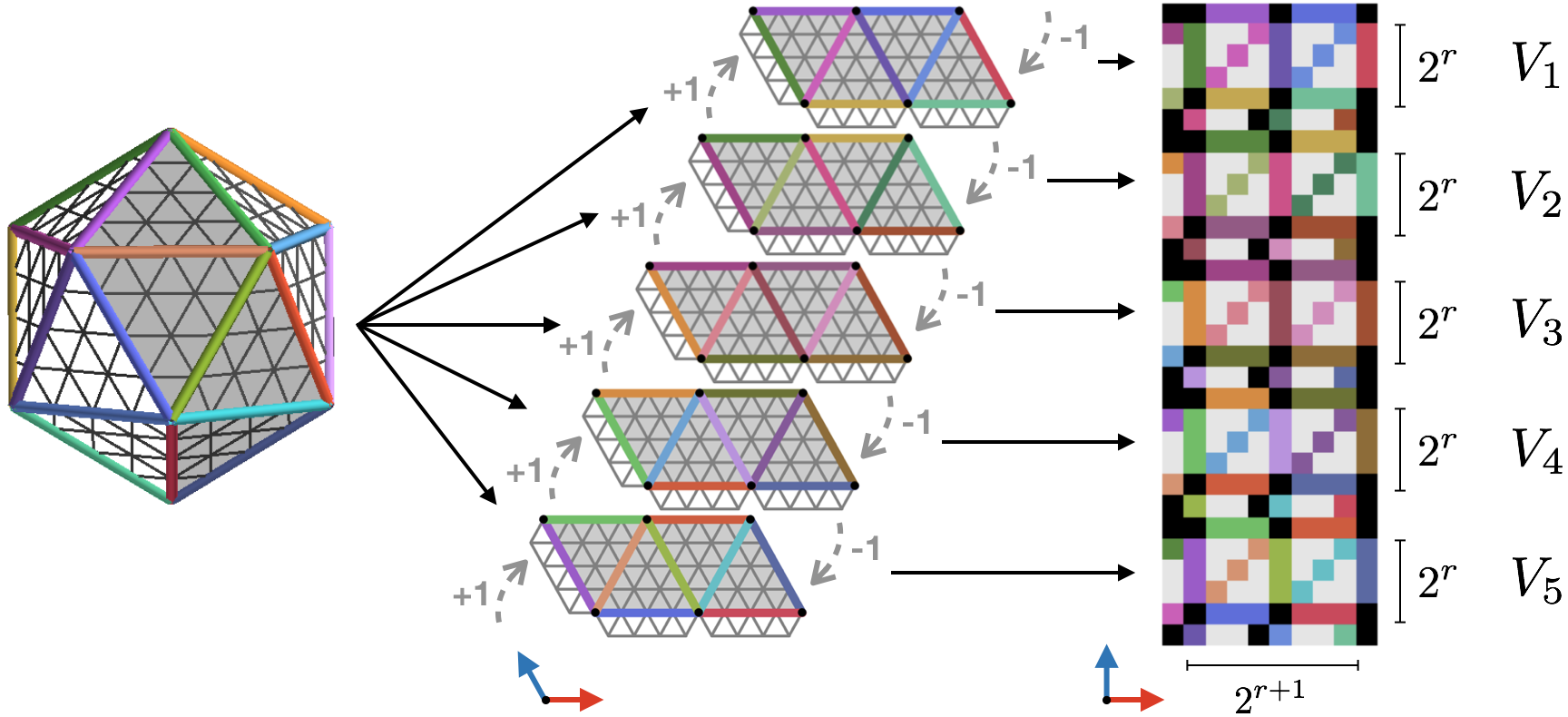}
    \caption{The Icosahedron with grid $\mathcal{H}_r$ for $r=2$ (left). We define $5$ overlapping charts that cover the grid (center). Chart $V_5$ is highlighted in gray (left). Colored edges that appear in multiple charts are to be identified. In each chart, we define the gauge by the standard axis aligned basis vectors $e_1, e_2 \in V_i$. For points $p \in U_i \cap U_j$, the transition between charts involves a change of gauge, shown as $+1 \cdot 2\pi/6$ and $-1 \cdot 2\pi /6$ (elements of $G=C_6$).
    On the right we show how the signal is represented in a padded array of shape $5 \cdot (2^r + 2) \times (2^{r+1}+2)$.}
    \label{fig:atlas}
\end{figure}

We define an \emph{atlas} consisting of $5$ overlapping \emph{charts} on the icosahedron, as shown in Fig. \ref{fig:atlas}.
Each chart is an invertible map $\varphi_i : U_i \rightarrow V_i$, where $U_i \subset \mathcal{H}_r \subset M$ and $V_i \subset \Z^2$.
The regions $U_i$ and $V_i$ are shown in Fig. \ref{fig:atlas}.
The maps themselves are linear on faces, and defined by hard-coded correspondences $\varphi_i(c_j) = x_j$ between the corner points $c_j$ in $\mathcal{H}_r$ and points $x_j$ in the planar grid $\Z^2$.

Each chart covers all the points in $4$ triangular faces of the icosahedron.
Together, the $5$ charts cover all $20$ faces of the icosahedron.

We divide the charts into an \emph{exterior} $\overline{V}_i \subset V_i$, consisting of border pixels, and an \emph{interior} $V^\circ_i \subset V_i$, consisting of pixels whose neighbours are all contained in chart $i$.
In order to ensure that every pixel in $\mathcal{H}_r$ (except for the $12$ corners) is contained in the interior of some chart, we add a strip of pixels to the left and bottom of each chart, as shown in Fig. \ref{fig:atlas} (center).
Then the interior of each chart (plus two exterior corners) has a nice rectangular shape $2^r \times 2^{r+1}$, and every non-corner is contained in exactly one interior $V^\circ_i$.

So if we know the values of the field in the interior of each chart, we know the whole field (except for the corners, which we ignore).
However, in order to compute a valid convolution output at each interior pixel (assuming a hexagonal filter with one ring, i.e. a $3 \times 3$ masked filter), we will still need the exterior pixels to be filled in as well (introducing a small amount of redundancy).
See Sec. \ref{sec:gauge_padding}.

\subsection{The Gauge}
\label{sec:gauge}

For the purpose of computation, we fix a convenient gauge in each chart. 
This gauge is defined in each $V_i$ as the constant orthogonal frame $e_1 = (1,0), e_2 = (0, 1)$, aligned with the $x$ and $y$ direction of the plane (just like the red and blue gauge in Fig. \ref{fig:gauge_trafo}).
When mapped to the icosahedron via (the Jacobian of) $\varphi_i^{-1}$, the resulting frames are aligned with the grid, and the basis vectors make an angle of $2 \cdot 2\pi / 6$.

Some pixels $p \in U_i \cap U_j$ are covered by multiple charts.
Although the local frames $e_1 = (1,0), e_2=(0,1)$ are numerically constant and equal in both charts $V_i$ and $V_j$, the corresponding frames on the icosahedron (obtained by pushing them though $\varphi_i^{-1}$ and $\varphi_j^{-1}$) may not be the same.
In other words, when switching from chart $i$ to chart $j$, there may be a gauge transformation $g_{ij}(p)$, which rotates the frame at $p \in U_i \cap U_j$ (see Fig. \ref{fig:gauge_trafo}).

For the particular atlas defined in Sec. \ref{sec:atlas}, the gauge transformations $g_{ij}(p)$ are always elements of the group $C_6$ (i.e. rotations by $k \cdot 2 \pi / 6$), so $G = C_6$ and we have a $C_6$-atlas.

\subsection{The Signal Representation}
\label{sec:signal_representation}

A stack of $C$ feature fields is represented as an array of shape $(B, C, R, 5, H, W)$, where $B$ is the batch size, $C$ the number of fields, $R$ is the dimension of the fields ($R=1$ for scalars and $R=6$ for regular features), $5$ is the number of charts, and $H, W$ are the height and width of each local chart ($H=2^r + 2$ and $W=2^{r+1} + 2$ at resolution $r$, including a $1$-pixel padding region on each side, see Fig. \ref{fig:atlas}).
We can always reshape such an array to shape $(B, CR, 5H, W)$, resulting in a $4D$ array that can be viewed as a stack of $C R$ rectangular feature maps of shape $(5H, W)$.
Such an array can be input to conv2d.

\subsection{Gauge Equivariant Icosahedral Convolution}

Gauge equivariant convolution on the icosahedron is implemented in three steps: G-Padding, kernel expansion, and 2d convolution / HexaConv \cite{hoogeboomHexaConv2018}.

\subsubsection{G-Padding}
\label{sec:gauge_padding}

\begin{wrapfigure}{r}{0.2\textwidth} 
    \centering
    \includegraphics[width=2cm]{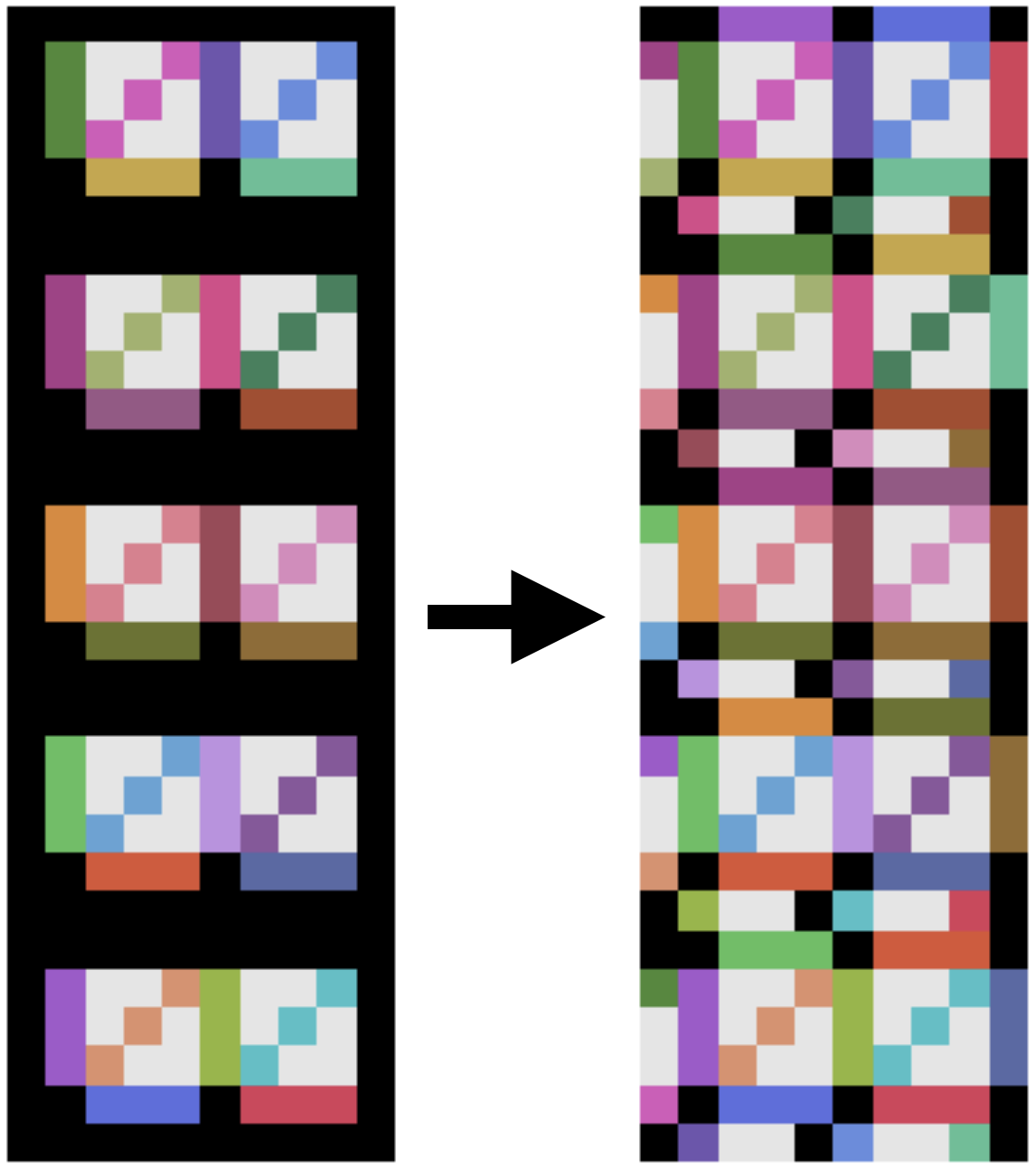}
    \caption{G-Padding (scalar signal)}
    \label{fig:gpad}
\end{wrapfigure}
In a standard CNN, we can only compute a valid convolution output at positions $(x, y)$ where the filter fits inside the input image in its entirety.
If the output is to be of the same size as the input, one uses zero padding.
Likewise, the IcoConv requires padding, only now the padding border $\overline{V}_i$ of a chart consists of pixels that are also represented in the interior of another chart (Sec. \ref{sec:atlas}).
So instead of zero padding, we copy the pixels from the neighbouring chart.
We always use hexagonal filters with 1 ring, which can be represented as a $3\times 3$ filter on a square grid, so we pad by 1 pixel.

As explained in Sec. \ref{sec:gauge}, when transitioning between charts one may have to perform a gauge transformation on the features.
Since scalars are invariant quantities, transition padding amounts to a simple copy in this case.
Regular $C_6$ features (having $6$ orientation channels) transform by cyclic shifts $\rho(g_{ij}(p))$ (Sec. \ref{sec:feature_fields}), where $g_{ij} \in \{+1, 0, -1\} \cdot 2\pi/6$ (Fig. \ref{fig:atlas}), so we must cyclically shift the channels up or down before copying to get the correct coefficients in the new chart.
The whole padding operation is implemented by four indexing + assignment operations (top, bottom, left, right) using fixed pre-computed indices (see Supp. Mat.).

\subsubsection{Weight Sharing \& Kernel Expansion}
\label{sec:kernel_expansion}

\begin{figure}[h!]
    \centering
    \includegraphics[width=2.5cm]{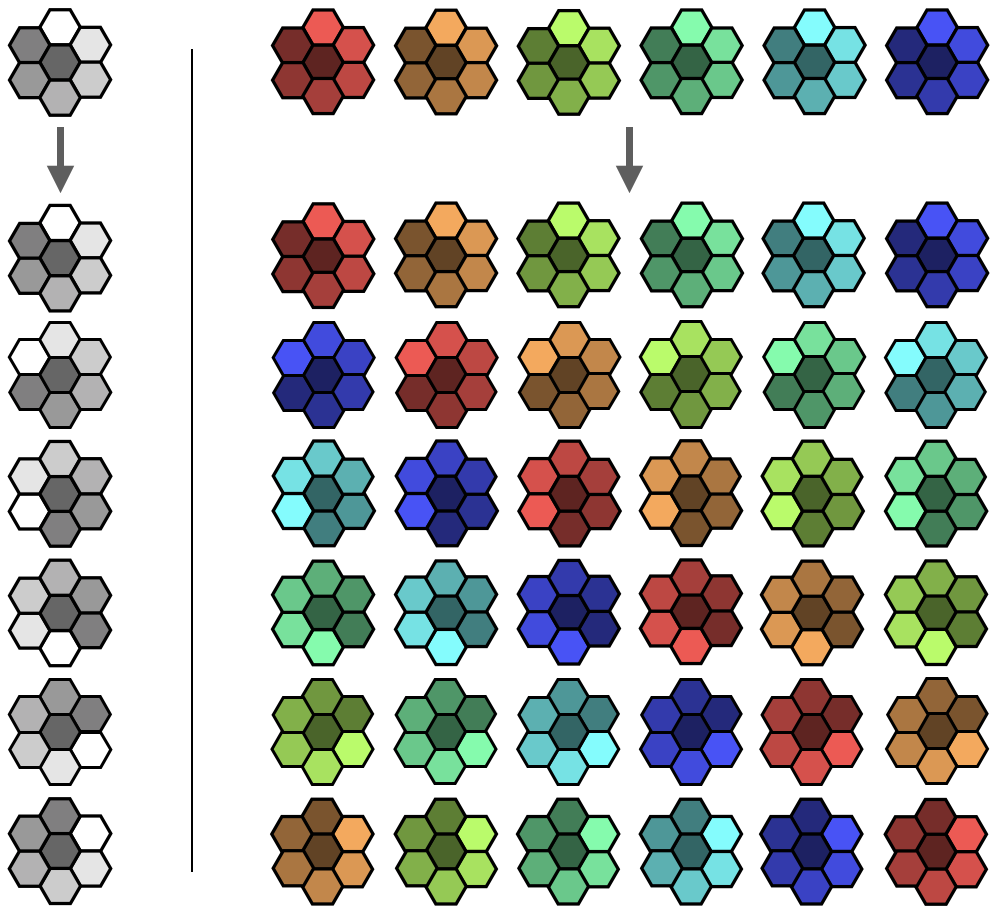}
    \caption{Kernel expansion for scalar-to-regular ($R_\text{in}=1, R_\text{out}=6$; left) and regular-to-regular ($R_\text{in} = R_\text{out}=6$; right) convolution. Top: free parameters. Bottom: expanded kernel used in conv2d.}
    \label{fig:kexp}
\end{figure}
For the convolution to be gauge equivariant, the kernel must satisfy Eq. \ref{eq:kernel_constraint}.
The kernel $K : \R^2 \rightarrow \R^{R_\text{out} C_\text{out} \times R_\text{in} C_\text{in}}$ is stored in an array of shape $(R_\text{out} C_\text{out}, R_\text{in} C_\text{in}, 3, 3)$, with the top-right and bottom-left pixel of each $3 \times 3$ filter fixed at zero so that it corresponds to a 1-ring hexagonal kernel.

Eq. \ref{eq:kernel_constraint} says that if we transform the input channels (columns) by $\rho_\text{in}(g)$ and the ouput channels (rows) by $\rho_\text{out}(g)$, the result should equal the original kernel where each channel is rotated by $g \in C_6$.
This will the case if we use the weight-sharing scheme shown in Fig. \ref{fig:kexp}.

Weight sharing can be implemented in two ways.
One can construct a basis of kernels, each of which has shape $(R_\text{out}, R_\text{in}, 3,3)$ and has value $1$ at all pixels of a certain color/shade, and $0$ elsewhere.
Then one can construct the full kernel by linearly combining these basis filters using learned weights (one for each $C_{\text{in}}\cdot C_{\text{out}}$ input/output channels and basis kernel) \cite{cohenSteerableCNNs2017, weiler3DSteerableCNNs2018}.
Alternatively, for scalar and regular features, one can use a set of precomputed indices to expand the kernel as shown in Fig. \ref{fig:kexp}, using a single indexing operation.

\subsubsection{Complete Algorithm}

The complete algorithm can be summarized as
\begin{equation}
     \text{GConv(f, w) = conv2d(GPad(f), expand(w))}.
\end{equation}
Where $f$ and $\text{GPad}(f)$ both have shape $(B, C_{\text{in}} R_\text{in}, 5H, W)$, the weights $w$ have shape $(C_{\text{out}}, C_{\text{in}} R_{\text{in}}, 7)$, and $\text{expand}(w)$ has shape $(C_{\text{out}} R_{\text{out}}, C_{\text{in}} R_{\text{in}}, 3, 3)$.
The output of GConv has shape $(B, C_{\text{out}} R_{\text{out}}, 5H, W)$.

On the flat faces, being described by one of the charts, this algorithm coincides exactly with the hexagonal regular convolution introduced in \cite{hoogeboomHexaConv2018}.
The non-zero curvature of the icosahedron requires us to do the additional step of padding between different charts.

\section{Experiments}
\label{sec:experiments}

\subsection{IcoMNIST}

In order to validate our implementation, highlight the potential benefits of our method, and determine the necessity of each part of the algorithm, we perform a number of experiments with the MNIST dataset, projected to the icosahedron.

We generate three different versions of the training and test sets, differing in the transformations applied to the data.
In the N condition, No rotations are applied to the data.
In the I condition, we apply all $60$ Icosahedral symmetries (rotations) to each digit.
Finally, in the R condition, we apply $60$ random continuous rotations $g \in \SO3$ to each digit before projecting.
All signals are represented as explained in Sec. \ref{sec:signal_representation} / Fig. \ref{fig:atlas} (right), using resolution $r=4$, i.e. as an array of shape $(1, 5 \cdot (16 + 2), 32 + 2)$.

Our main model consists of one gauge equivariant scalar-to-regular (S2R) convolution layer, followed by $6$ regular-to-regular (R2R) layers and $3$ FC layers (see Supp. Mat. for architectural details).
We also evaluate a model that uses only S2R convolution layers, followed by orientation pooling (a $\max$ over the $6$ orientation channels of each regular feature, thus mapping a regular feature to a scalar), as in \cite{masciGeodesicConvolutionalNeural2015}.
Finally, we consider a model that uses only rotation-invariant filters, i.e. scalar-to-scalar (S2S) convolutions, similar to standard graph CNNs \cite{boscainiLearningClassSpecific2015, kipfSemiSupervisedClassificationGraph2017}.
We also compare to the fully $\SO3$-equivariant but computationally costly Spherical CNN (S2CNN).
See supp. mat. for architectural details and computational complexity analysis.

In addition, we perform an ablation study where we disable each part of the algorithm.
The first baseline is obtained from the full R2R network by disabling gauge padding (Sec. \ref{sec:gauge_padding}), and is called the No Pad (NP) network.
In the second baseline, we disable the kernel Expansion (Sec. \ref{sec:kernel_expansion}), yielding the NE condition.
The third baseline, called NP+NE uses neither gauge padding nor kernel expansion, and amounts to a standard CNN applied to the same input representation.
We adapt the number of channels so that all networks have roughly the same number of parameters.

\begin{table}[b]
\small
\centering
\begin{tabular}{c | c c c c c c}
    Arch.  & N/N            & N/I & N/R             & I/ I & I / R & R / R \\
    \hline
    S2CNN & {\tiny $99.38$} & {\tiny $99.38$} & {\tiny $99.38$} & {\tiny $99.12$} & {\tiny $99.13$} & {\tiny $99.12$} \\
    \hline
    NP+NE & {\tiny $99.29$}  & {\tiny $25.50$} & {\tiny $16.20$} & {\tiny $98.52$} & {\tiny $47.77$} & {\tiny $94.19$} \\
    NE    & {\tiny $99.42$} & {\tiny $25.41$} & {\tiny $17.85$} & {\tiny $98.67$}            & {\tiny $60.74$}   & {\tiny $96.83$} \\
    NP    & {\tiny $99.27$} & {\tiny $36.76$} & {\tiny $21.4$} & {\tiny $98.99$} & {\tiny $61.62$} & {\tiny $97.87$} \\
    S2S   & {\tiny $97.81$} & {\tiny $97.81$} & {\tiny $55.64$} & {\tiny $97.72$}            & {\tiny $58.37$}   & {\tiny $89.92$} \\
    S2R   & {\tiny $98.99$} & {\tiny $98.99$} & {\tiny $59.76$} & {\tiny $98.62$}            & {\tiny $55.57$}   & {\tiny $98.74$} \\
    R2R   & {\tiny $\mathbf{99.43}$} & {\tiny $\mathbf{99.43}$} & {\tiny $\mathbf{69.99}$} & {\tiny $\mathbf{99.38}$}            & {\tiny $\mathbf{66.26}$} & {\tiny $\mathbf{99.31}$} \\
\end{tabular}
\caption{IcoMNIST test accuracy (\%) for different architectures and train / test conditions (averaged over 3 runs). See text for explanation of labels.}
\label{tab:mnist}
\end{table}

As shown in Table \ref{tab:mnist}, icosahedral CNNs achieve excellent performance with a test accuracy of up to 99.43\%, which is a strong result even on planar MNIST, for non-augmented and non-ensembled models.
The full R2R model performs best in all conditions (though not significantly in the N/N condition), showing that both gauge padding and kernel expansion are necessary, and that our general (R2R) formulation works better in practice than using scalar fields (S2S or S2R).
We notice also that non-equivariant models (NP+NE, NP, NE) do not generalize well to transformed data, a problem that is only partly solved by data augmentation.
On the other hand, the models S2S, S2R, and R2R are exactly equivariant to symmetries $g \in \mathcal{I}$, and so generalize perfectly to $\mathcal{I}$-transformed test data, even when these were not seen during training.
None of the models automatically generalize to continuously rotated inputs (R), but the equivariant models are closer, and can get even closer ($>99\%$) when using $\SO3$ data augmentation during training.
The fully $\SO3$-equivariant S2CNN scores slightly worse than R2R, except in N/R and I/R, as expected.
The slight decrease in performance of S2CNN for rotated training conditions is likely due to the fact that it has lower grid resolution near the equator.
We note that the S2CNN is slower and less scalable than Ico CNNs (see supp. mat.).

\subsection{Climate Pattern Segmentation}

We evaluate our method on the climate pattern segmentation task proposed by \citet{mudigondaSegmentingTrackingExtreme}.
The goal is to segment extreme weather events (Atmospheric Rivers (AR) and Tropical Cyclones (TC)) in climate simulation data.

We use the exact same data and evaluation methodology as \cite{jiangSphericalCNNsUnstructured2018}.
The preprocessed data as released by \cite{jiangSphericalCNNsUnstructured2018} consists of $16$-channel spherical images at resolution $r=5$, which we reinterpret as icosahedral signals (introducing slight distortion).
See \cite{mudigondaSegmentingTrackingExtreme} for a detailed description of the data.

We compare an R2R and S2R model (details in Supp. Mat.).
As shown in Table \ref{tab:climate}, our models outperform both competing methods in terms of per-class and mean accuracy.
The difference between our R2R and S2R model seems small in terms of accuracy, but when evaluated in terms of mean average precision (a more appropriate evaluation metric for segmentation tasks), the R2R model clearly outperforms.

\begin{table}[h!]
    \centering
    \begin{tabular}{l c c c c c}
        \toprule
        Model & BG & TC & AR & Mean & mAP \\
        \midrule
        {\small Mudigonda et al.}
         & 97 & 74 & 65 & 78.67 & - \\
        {\small Jiang et al.}
         & 97 & 94 & 93 & 94.67 & - \\
        Ours (S2R) & 97.3 & 97.8 & 97.3 & 97.5 & 0.686 \\
        Ours (R2R) & 97.4 & \textbf{97.9} & \textbf{97.8} & \textbf{97.7} & \textbf{0.759} \\
        \bottomrule
    \end{tabular}
    \caption{Climate pattern segmentation accuracy (\%) for BG, TC and AR classes plus mean accuracy and average precision (mAP).}
    \label{tab:climate}
\end{table}

\subsection{Stanford 2D-3D-S}

For our final experiment, we evaluate icosahedral CNNs on the 2D-3D-S dataset \cite{armeniJoint2D3DSemanticData}, which consists of 1413 omnidirectional RGB+D images with pixelwise semantic labels in 13 classes. 
Following \citet{jiangSphericalCNNsUnstructured2018}, we sample the data on a grid of resolution $r=5$ using bilinear interpolation, while using nearest-neighbour interpolation for the labels.
Evaluation is performed by mean intersection over union (mIoU) and pixel accuracy (mAcc).

The network architecture is a residual U-Net \cite{RFB15a, heDeepResidualLearning2016} with R2R convolutions.
The network consists of a downsampling and upsampling network.
The downsampling network takes as input a signal at resolution $r = 5$ and outputs feature maps at resolutions $r=4, \ldots, 1$, with $8, 16, 32$ and $64$ channels.
The upsampling network is the reverse of this.
We pool over orientation channels right before applying softmax.

As shown in table \ref{tab:2d3ds}, our method outperforms the method of \cite{jiangSphericalCNNsUnstructured2018}, which in turn greatly outperforms standard planar methods such as U-Net on this dataset.
\begin{table}[h!]
    \centering
    \begin{tabular}{l c c}
        \toprule
         & mAcc & mIoU \\
        \midrule
        {\small \cite{jiangSphericalCNNsUnstructured2018}}
         & $0.547$ & $0.383$ \\
        Ours (R2R-U-Net) & $\mathbf{0.559}$ & $\mathbf{0.394}$ \\
        \bottomrule
    \end{tabular}
    \caption{Mean accuracy and intersection over union for 2D-3D-S omnidirectional segmentation task.}
    \label{tab:2d3ds}
\end{table}

\section{Conclusion}

In this paper we have presented the general theory of gauge equivariant convolutional networks on manifolds, and demonstrated their utility in a special case: learning with spherical signals using the icosahedral CNN.
We have demonstrated that this method performs well on a range of different problems and is highly scalable.

Although we have only touched on the connections to physics and geometry, there are indeed interesting connections, which we plan to elaborate on in the future.
From the perspective of the mathematical theory of principal fiber bundles, our definition of manifold convolution is entirely natural.
Indeed it is clear that gauge invariance is not just nice to have, but \emph{necessary} in order for the convolution to be geometrically well-defined.

In future work, we plan to implement gauge CNNs on general manifolds and work on further scaling of spherical CNNs.
Our chart-based approach to manifold convolution should in principle scale to very large problems, thus opening the door to learning from high-resolution planetary scale spherical signals that arise in the earth and climate sciences.

\section*{Acknowledgements}
We would like to thank Chiyu ``Max'' Jiang and Mayur Mudigonda for help obtaining and interpreting the climate data, and Erik Verlinde for helpful discussions.
\blfootnote{The climate dataset released by \cite{jiangSphericalCNNsUnstructured2018} and the Stanford 2D-3D-S datasets were downloaded and evaluated by QUvA researchers.}

\bibliography{main}
\bibliographystyle{icml2019}

\newpage
\part*{Supplementary Material}
\input{supp_content}

\end{document}

%% file: supp_content.tex
\section{Recommended reading}

For more information on manifolds, fiber bundles, connections, parallel transport, the exponential map, etc., we highly recommend the lectures by \citet{schullerLecturesGeometricalAnatomy2016}, as well as the book \citet{nakaharaGeometryTopologyPhysics2003} which explain these concepts very clearly and at a useful level of abstraction.

For further study, we recommend \cite{sharpeDifferentialGeometryCartan1997, shoshichikobayashiFoundationsDifferentialGeometry1963, husemollerFibreBundles1994a, steenrodTopologyFibreBundles, wendlLectureNotesBundles2008, craneDiscreteDifferentialGeometry2014}.

\section{Mathematical Theory \& Physics Analogy}
\label{sec:theory}

From the perspective of the theory of principal fiber bundles, our work can be understood as follows.
A fiber bundle $E$ is a space consisting of a base space $B$ (the manifold $M$ in our paper), with at each point $p \in B$ a space $F_p$ called the fiber at $p$.
The bundle is defined in terms of a projection map $\pi : E \rightarrow B$, which determines the fibers as $F_p = \pi^{-1}(p)$.
A principal bundle is a fiber bundle where the fiber $F$ carries a transitive and free right action of a group $G$ (the structure group).

One can think of the fiber $F_p$ of a principal bundle as a (generalized) space of frames at $p$.
Due to the free and transitive action of $G$ on $F_p$, we have that $F_p$ is isomoprhic to $G$ as a $G$-space, meaning that it looks like $G$ except that it does not have a distinguished origin or identity element as $G$ does (i.e. there is no natural choice of frame).

A gauge transformation is then defined as a principal bundle automorphism, i.e. a map from $P \rightarrow P$ that maps fibers to fibers in a $G$-equivariant manner.
Sometimes the automorphism is required to fix the base space, i.e. project down to the identity map via $\pi$.
Such a $B$-automorphism will map each fiber onto itself, so it restricts to a $G$-space automorphism on each fiber.

Given a principal bundle $P$ and a vector space $V$ with representation $\rho$ of $G$, we can construct the associated bundle $P \times_\rho V$, whose elements are the equivalence classes of the following equivalence relation on $P \times V$:
\begin{equation}
    (p, v) \sim (pg, \rho(g^{-1})v).
\end{equation}
The associated bundle is a fiber bundle over the same base space as $P$, with fiber isomorphic to $V$.

A (matter) field is described as a section of the associated bundle $A$, i.e. a map 
$\sigma : B \rightarrow A$ that satisfies $\pi \circ \sigma = 1_B$.
Locally, one can describe a section as a function $B \rightarrow V$ (as we do in the paper), but globally this is not possible unless the bundle is trivial.

The group of automorphisms of $P$ (gauge transformations) acts on the space of fields (sections of the associated bundle).
It is this group that we wish to be equivariant to.

From this mathematical perspective, our work amounts to replacing the principal $G$ bundle\footnote{It is more common to use the letter $G$ for the supergroup and $H$ for the subgroup, but that leads to a principal $H$-bundle $G \rightarrow G/H$, which is inconsistent with the main text, where we use a principal $G$ bundle. So we swap $H$ and $G$ here.} $H \rightarrow H/G$ used in the work on regular and steerable G-CNNs of \citet{cohenGeneralTheoryEquivariant2018,cohenIntertwinersInducedRepresentations2018}, by another principal $G$ bundle, namely the frame bundle of $M$.
Hence, this general theory can describe in a unified way the most prominent and geometrically natural methods of geometrical deep learning \cite{masciGeodesicConvolutionalNeural2015, boscainiLearningShapeCorrespondence2016}, as well as all G-CNNs on homogeneous spaces.

Indeed, if we build a gauge equivariant CNN on a homogeneous space $H/G$ (e.g. the sphere $S^2 = \operatorname{SO}(3) / \operatorname{SO}(2)$), it will (under mild conditions) automatically be equivariant to the left action of $H$ also.
To see this, note that the left action of $H$ on itself (the total space of the principal $G$ bundle) can be decomposed into an action on the base space $H/G$ (permuting the fibers), and an action on the fibers (cosets) that factors through $G$ (see e.g. Sec. 2.1 of \cite{cohenIntertwinersInducedRepresentations2018}).
The action on the base space preserves the local neighbourhoods from which we compute filter responses, and equivariance to the action of $G$ is ensured by the kernel constraint. 
Since G-CNNs \cite{cohenGeneralTheoryEquivariant2018} and gauge equivariant CNNs employ the most general equivariant map, we conclude that they are indeed the same, for bundles $H \rightarrow H/G$.
Thus, ``gauge theory is all you need''.
(We plan to expand this argument in a future paper)

Most modern theories of physics are gauge theories, meaning they are based on this mathematical framework.
In such theories, any construction is required to be gauge invariant (i.e. the coefficients must be gauge equivariant), for otherwise the predictions will depend on the way in which we choose to represent physical quantities.
This logic applies not just to physics theories, but, as we have argued in the paper, also to neural networks and other models used in machine learning.
Hence, it is only natural that the same mathematical framework is applicable in both fields.

\section{Deriving the kernel constraint}

The gauge equivariant convolution is given by
\begin{equation}
    (K \star f)(p) = \int_{\R^d} K(v) \rho_\text{in}(g_{p \leftarrow q_v} ) f(q_v) dv.
\end{equation}

Under a gauge transformation, we have:
\begin{equation}
    \begin{aligned}
        v &\mapsto g_p^{-1} v, \;\;\;\;\; & f(q_v) &\mapsto \rho_\text{in}(g_{q_v}^{-1}) f(q_v), \\
        w_p & \mapsto w_p g_p, \;\;\;\;\; & g_{p \leftarrow q_v} &\mapsto g_p^{-1} g_{p \leftarrow q_v} g_{q_v}.
    \end{aligned}
\end{equation}

It follows that $q_v$ is unchanged, because $q_v = \exp_p w_p v \mapsto \exp_p (w_p g_p) (g_p^{-1} v) = q_v$.
Substituting the rest in the convolution equation, we find
\begin{equation}
    \begin{aligned}
        \int_{\R^d} K(g_p^{-1} v) \rho_\text{in}(g_p^{-1} g_{p \leftarrow q_v } g_{q_v}) \rho_\text{in}(g_{q_v}^{-1}) f(q_v ) dv \\
        = 
        \int_{\R^d} K(g_p^{-1} v) \rho_\text{in}(g_p^{-1}) \rho_\text{in}(g_{p \leftarrow q_v }) f(q_v ) dv
    \end{aligned}
\end{equation}
Now if $K(g_p^{-1} v) = \rho_\text{out}(g_p^{-1}) K(v) \rho_\text{in}(g_p)$ (i.e. $K$ satisfies the kernel constraint), then we get
\begin{equation}
    (K \star f)(p) \mapsto \rho_\text{out}(g_p^{-1}) (K \star f)(p), 
\end{equation}
i.e. $K \star f$ transforms as a $\rho_\text{out}$-field under gauge transformations.

\section{Additional information on experiments}

\subsection{MNIST experiments}

Our main model consists of $7$ convolution layers and $3$ linear layers.
The first layer is a scalar-to-regular gauge equivariant convolution layer, and the following $6$ layers are regular-to-regular layers.
These layers have $8, 16, 16, 24, 24, 32, 64$ output channels, and stride $1, 2, 1, 2, 1, 2, 1$, respectively.

In between convolution layers, we use batch normalization \cite{ioffeBatchNormalizationAccelerating2015} and ReLU nonlinearities.
When using batch normalization, we average over groups of $6$ feature maps, to make sure the operation is equivariant.
Any pointwise nonlinearity (like ReLU) is equivariant, because we use only trivial and regular representations realized by permutation matrices.

After the convolution layers, we perform global pooling over spatial and orientation channels, yielding an invariant representation.
We map these through 3 FC layers (with $64, 32, 10$ channels) before applying softmax.

The other models are obtained from this one by replacing the convolution layers by scalar-to-regular + orientation pooling (S2R) or scalar-to-scalar (S2S) layers, or by disabling G-padding (NP) and/or kernel expansion (NE), always adjusting the number of channels to keep the number of parameters roughly the same.

The Spherical CNN (S2CNN) is obtained from the R2R model by replacing the S2R and R2R layers by spherical and $\SO3$ convolution layers, respectively, keeping the number of channels and strides the same.
The Spherical CNN uses a different grid than the Icosahedral CNN, so we adapt the resolution / bandwidth parameter $B$ to roughly match the resolution of the Icosahedral CNN.
We use $B = 26$, to get a spherical grid of size $2B \times 2B = 52 \times 52$.
Note that this grid has higher resolution at the poles, and lower resolution near the equator, which explains why the S2CNN performs a bit worse when trained on rotated data instead of digits projected onto the north-pole.
To implement strides, we reduce the output bandwidth by $2$ at each layer with stride.

The spherical convolution takes a scalar signal on the sphere as input, and outputs scalar signals on $\SO3$, which is analogous to a regular field over the sphere.
$\SO3$ convolutions are analogous to $R2R$ layers.
We note that this is a stronger Spherical CNN architecture than the one used by \cite{cohenSphericalCNNs2018}, which achieves only $96\%$ accuracy on spherical MNIST.

The models were trained for $60$ epochs, or $1$ epoch of the $60\times$ augmented dataset (where each instance is transformed by each icosahedron symmetry $g \in \mathcal{I}$, or by a random rotation $g \in \operatorname{SO}(3)$).

\subsection{Climate experiments}

For the climate experiments, we used a U-net with regular-to-regular convolutions. 
The first layer is a scalar-to-regular convolution with 16 output channels.
The downsampling path consists of $5$ regular-to-regular layers with stride $2$, and $32, 64, 128, 256, 256$ output channels.
The downsampling path takes as input a signal with resolution $r = 5$ (i.e. $10242$ pixels), and outputs one at $r = 0$ (i.e. $12$ pixels).

The decoder is the reverse of the encoder in terms of resolution and number of channels.
Upsampling is performed by bilinear interpolation (which is exactly equivariant), before each convolution layer (which uses stride 1).
As usual in the U-net architecture, each layer in the upsampling path takes as input the output of the previous layer, as well as the output of the encoder path at the same resolution.

Each convolution layer is followed by equivariant batchnorm and ReLU.

The model was trained for $15$ epochs with batchsize $15$.

\subsection{2D-3D-S experiments}

For the 2D-3D-S experiments, we used a residual U-Net with the following architecture.

The input layer is a scalar-to-regular layer with 8 channels, followed by batchnorm and relu.
Then we apply 4 residual blocks with 16, 32, 64, 64 output channels, each of which uses stride=2.
In the upsampling stream, we use 32, 16, 8, 8 channels, for the residual blocks, respectively.
Each upsampling layer receives input from the corresponding downsampling layer, as well as the previous layer.
Upsampling is performed using bilinear interpolation, and downsampling by hexagonal max pooling.

The input resolution is $r=5$, which is downsampled to $r=1$ by the downsampling stream.

Each residual block consists of a convolution, batchnorm, skipconnection, and ReLU.

\section{Computational complexity analysis of Spherical and Icosahedral CNNs}

One of the primary motivations for the development of the Icosahedral CNN is that it is faster and more scalable than Spherical CNNs as originally proposed.
The Spherical CNN as implemented by \cite{cohenSphericalCNNs2018} uses feature maps on the sphere $S^2$ and rotation group $\SO3$ (the latter of which can be thought of a regular field on the sphere), sampled on the SOFT grids defined by \cite{kostelecSOFTFourierTransforms2007}, which have shape $2B \times 2B$ and $2B \times 2B \times 2B$, respectively (here $B$ is the bandwidth / resolution parameter).
Specifically, the grid points are:
\begin{equation}
    \begin{aligned}
        \alpha_{j_1} &= \frac{2\pi j_1}{2B}, \\
        \beta_{k} &= \frac{\pi(2k + 1)}{4B}, \\
        \gamma_{j_2} &= \frac{2\pi j_2}{2B}, \\
    \end{aligned}
\end{equation}
where $(\alpha_{j_1}, \beta_k)$ form a spherical grid and $(\alpha_{j_1}, \beta_k, \gamma_{j_2})$ form an $\SO3$ grid (for $j_1, k, j_2 = 0, \ldots 2B - 1$).
These grids have two downsides.

Firstly, because the SOFT grid consists of equal-lattitude rings with a fixed number of points (2B), the spatial density of points is inhomogeneous, with a higher concentration of points near the poles.
To get a sufficiently high sampling near the equator, we are forced to oversample the poles, and thus waste computational resources.
For almost all applications, a more homogeneous grid is more suitable.

The second downside of the SOFT grid on $\SO3$ is that the spatial resolution ($2B \times 2B$; $\alpha, \beta$) and angular resolution ($2B$; $\gamma$) are both coupled to the same resolution / bandwidth parameter $B$.
Thus, as we increase the resolution of the spherical image, the number of rotations applied to each filter is increased as well, which is undesirable.

The grid used in the Icosahedral CNN addresses both concerns.
It is spatially very homogeneous, and we apply the filters in $6$ orientations, regardless of spatial resolution.

The generalized FFT algorithm used by \cite{cohenGeneralTheoryEquivariant2018} only works on the SOFT grid.
Generalized FFTs for other grids exist \cite{kunisFastSphericalFourier2003}, but are harder to implement.
Moreover, although the (generalized) FFT can improve the asymptotic complexity of convolution for large input signals, the FFT-based convolution actually has worse complexity if we assume a fixed filter size.
That is, the $\SO3$ convolution (used in most layers of a typical Spherical CNN) has complexity $O(B^3 \log B)$ which compares favorably to the naive $O(B^6)$ spatial implementation.
However, if we use filters with a fixed (and usually small) size, the complexity of a naive spatial implementation reduces to $O(B^3)$, which is slightly better than the FFT-based implementation.
Furthermore, because the Icosahedral CNN uses a fixed number of orientations per filter (i.e. $6$), its computational complexity is even better: it is linear in the number of pixels of the grid, and so comparable to $O(B^2)$ for the SOFT grid.

\begin{figure}[h!]
    \centering
    \includegraphics[width=8cm]{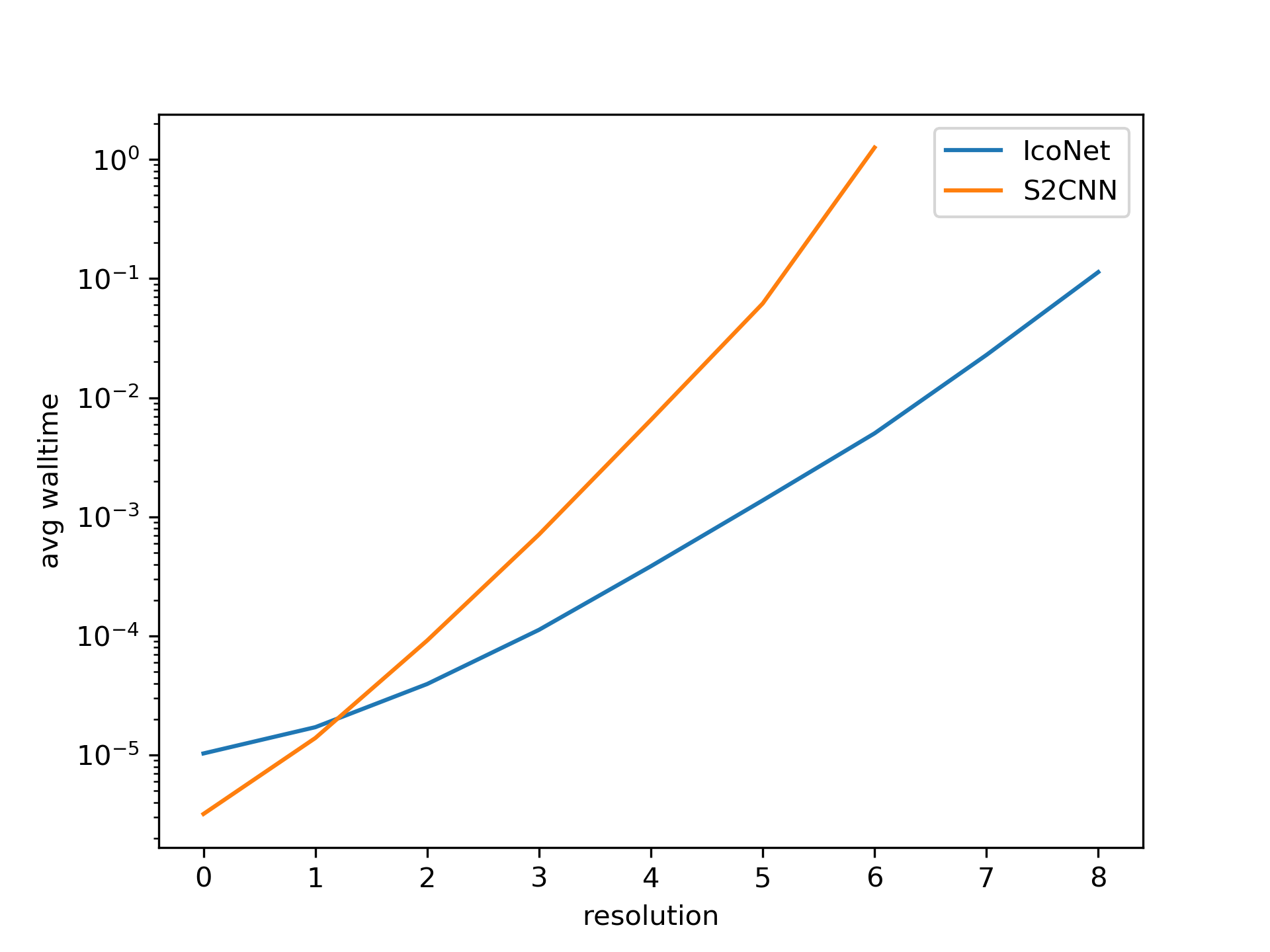}
    \caption{Comparison of computational cost (in wallclock time) of Icosahedral CNNs (IcoNet) and Spherical CNNs (S2CNN, \cite{cohenSphericalCNNs2018}), at increasing grid resolution $r$.}
    \label{fig:compute_comparison}
\end{figure}

\begin{figure}[h!]
    \centering
    \includegraphics[width=8cm]{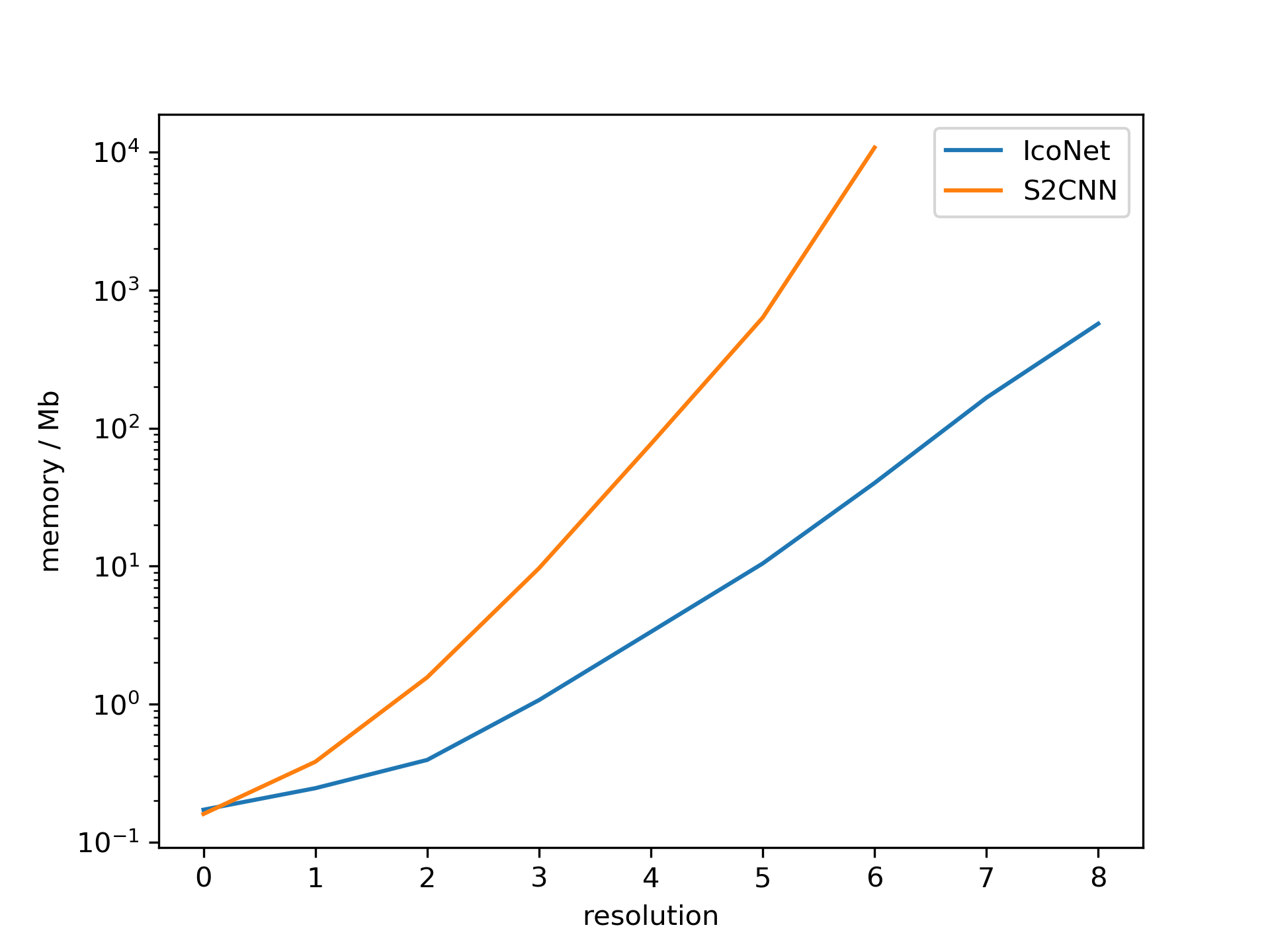}
    \caption{Comparison of memory usage of Icosahedral CNNs (IcoNet) and Spherical CNNs (S2CNN, \cite{cohenSphericalCNNs2018}), at increasing grid resolution $r$.}
    \label{fig:memory_comparison}
\end{figure}

The difference in complexity is clearly visible in Figures \ref{fig:compute_comparison} and \ref{fig:memory_comparison}, below.
On the horizonal axis, we show the grid resolution $r$ for the icosahedral grid $\mathcal{H}_r$ (for the spherical CNN, we a SOFT grid with roughly the same number of spatial points).
On the vertical axis, we show the amount of wallclock time (averaged over 100 runs) and memory required to run an $\SO3$ convolution (S2CNN) or a regular-to-regular gauge equivariant convolution (IcoNet) at that resolution.
Note that since the number of grid pionts is exponential in $r$, and we use a logarithmic vertical axis, the figures can be considered log-log plots.
Both plots were generated by running a single regular to regular convolution layer at the corresponding resolution $r$ with $12$ input and output channels.
For a fair comparison with IcoCNNs we chose filter grid parameters so3\_near\_identity\_grid(n\_alpha=6, max\_beta=np.pi/16, n\_beta=1, max\_gamma=2*np.pi, n\_gamma=6) for the spherical convolution layer.
To guarantee a full GPU utilization, results were measured on an as large as possible batch size per datapoint and subsequently normalized by that batch size.

As can be seen in Figure \ref{fig:compute_comparison}, the computational cost of running the S2CNN dramatically exceeds the cost of running the IcoCNN, particularly at higher resolutions.
We did not run the spherical CNN beyond resolution $r=6$, because the network would not fit in GPU memory even when using batch size $1$.

As shown in Figure \ref{fig:memory_comparison}, the Spherical CNN at resolution $r=6$ uses about 10GB of memory, whereas the Icosahedral CNN uses only about 1GB.
Since we used the maximum batch size with subsequent normalization for each resolution the reported memory consumption mainly reflects the memory cost of the feature maps, not the constant memory cost of the filter banks.

Aside from the theoretical asymptotic complexity, the actual computational cost depends on important implementation details.
Because the heavy lifting of the Icosahedral CNN is all done by a single conv2d call, our method benefits from the extensive optimization of, and hardware support for this operation.
By contrast, the generalized FFT used in the original Spherical CNN uses a conventional FFT, as well as matrix multiplcations with spectral matrices of size $1, 3, 5, 7, \ldots, 2 L + 1$ (the $\SO3$ spectrum is matrix-valued, instead of the scalar valued spectrum for commutative groups).
Implementing this in a way that is fast in practice is more challenging.

A final note on scalability.
For some problems, such as the analysis of high resolution global climate or weather data, it is unlikely that even a single feature map will fit in memory at once on current or near-term future hardware.
Hence, it may be useful to split large feature maps into local charts, and process each one on a separate compute node.
For the final results to be globally consistent (so that each compute node makes equivalent predictions for points in the overlap of charts), gauge equivariance is indispensable.

\section{Details on G-Padding}

In a conventional CNN, one has to pad the input feature map in order to compute an output of the same size.
Although the icosahedron itself does not have a boundary, the charts do, and hence require padding before convolution.
However, in order to faithfully simulate convolution on the icosahedron via convolution in the charts, the padding values need to be copied from another chart instead of e.g. padding by zeros.
In doing so, a gauge transformation may be required.

To see why, note that the conv2d operation, which we use to perform the convolution in the charts, implicitly assumes that the signal is expressed relative to a fixed global gauge in the plane, namely the frame defined by the x and y axes.
This is because the filters are shifted along the x and y directions by conv2d, and as they are shifted they are not rotated.
So the meaning of ``right'' and ``up'' doesn't change as we move over the plane; the local gauge at each position is aligned with the global x and y axes.

Hence, it is this global gauge that we must use inside the charts shown in Figure 4 (right) of the main paper.
It is important to note that although all frames have the same numerical expression $e_1 = (1,0), e_2 = (0, 1)$ relative to the x and y axes, the corresponding frames on the icosahedron itself are different for different charts.
Since feature vectors are represented by coefficients that have a meaning only relative to a frame, they have a different numerical expression in different charts in which they are contained.
The numerical representations of a feature vector in two charts are related by a gauge transformation.

To better understand the gauge transformation intuitively, consider a pixel $p$ on a colored edge in Fig. 4 of the main paper, that lies in multiple charts. 
Now consider a vector attached at this pixel (i.e. in $T_p M$), pointing along the colored edge.
Since the colored edge may have different orientations when pictured in different charts, the vector (which is aligned with this edge) will also point in different directions in the charts, when the charts are placed on the plane together as in Figure 4.
More specifically, for the choice of charts we have made, the difference in orientation is always one ``click'', i.e. a rotation by plus or minus $2\pi/6$.
This is the gauge transformation $g_{ij}(p)$, which describes the transformation at $p$ when switching between chart $i$ and $j$.

The transformation $g_{ij}(p)$ acts on the feature vector at $p$ via the matrix $\rho(g_{ij}(p))$, where $\rho$ is the representation of $G = C_6$ associated with the feature space under consideration.
In this work we only consider two kinds of representations: scalar features with $\rho(g) = 1$, and regular features with $\rho$ equal to the regular representation:
\begin{equation}
    \rho(2\pi/6) =
    \begin{bmatrix}
        0 & 1 & 0 & 0 & 0 & 0 \\
        0 & 0 & 1 & 0 & 0 & 0 \\
        0 & 0 & 0 & 1 & 0 & 0 \\
        0 & 0 & 0 & 0 & 1 & 0 \\
        0 & 0 & 0 & 0 & 0 & 1 \\
        1 & 0 & 0 & 0 & 0 & 0 \\
    \end{bmatrix}.
\end{equation}
That is, a cyclic permutation of 6 elements.
Since $2\pi/6$ is a generator of $C_6$, the value of $\rho$ at the other group elements is determined by this matrix: $\rho(k \cdot 2 \pi / 6) = \rho(2 \pi / 6)^k$.
If the feature vector consists of multiple scalar or regular features, we would have a block-diagonal matrix $\rho(g_{ij}(p))$.

We implement G-padding by indexing operations on the feature maps.
For each position $p$ to be padded, we pre-compute $g_{ij}(p)$, which can be $+1 \cdot 2\pi/6$ or $0$ or $-1 \cdot 2 \pi / 6$.
We use these to precompute four indexing operations (for the top, bottom, left and right side of the charts).

%% file: main.bbl
\begin{thebibliography}{55}
\providecommand{\natexlab}[1]{#1}
\providecommand{\url}[1]{\texttt{#1}}
\expandafter\ifx\csname urlstyle\endcsname\relax
  \providecommand{\doi}[1]{doi: #1}\else
  \providecommand{\doi}{doi: \begingroup \urlstyle{rm}\Url}\fi

\bibitem[{Armeni} et~al.(2017){Armeni}, {Sax}, {Zamir}, and
  {Savarese}]{armeniJoint2D3DSemanticData}
{Armeni}, I., {Sax}, A., {Zamir}, A.~R., and {Savarese}, S.
\newblock {Joint 2D-3D-Semantic Data for Indoor Scene Understanding}.
\newblock \emph{ArXiv e-prints}, February 2017.

\bibitem[Bekkers et~al.(2018)Bekkers, Lafarge, Veta, Eppenhof, Pluim, and
  Duits]{bekkersRotoTranslationCovariantConvolutional2018}
Bekkers, E.~J., Lafarge, M.~W., Veta, M., Eppenhof, K. A.~J., Pluim, J. P.~W.,
  and Duits, R.
\newblock Roto-{Translation} {Covariant} {Convolutional} {Networks} for
  {Medical} {Image} {Analysis}.
\newblock In \emph{MICCAI}, 2018.

\bibitem[Boomsma \& Frellsen(2017)Boomsma and
  Frellsen]{boomsmaSphericalConvolutionsTheir2017}
Boomsma, W. and Frellsen, J.
\newblock Spherical convolutions and their application in molecular modelling.
\newblock In \emph{NIPS}, 2017.

\bibitem[Boscaini et~al.(2015)Boscaini, Masci, Melzi, Bronstein, Castellani,
  and Vandergheynst]{boscainiLearningClassSpecific2015}
Boscaini, D., Masci, J., Melzi, S., Bronstein, M.~M., Castellani, U., and
  Vandergheynst, P.
\newblock Learning class-specific descriptors for deformable shapes using
  localized spectral convolutional networks.
\newblock \emph{Computer Graphics Forum}, 2015.

\bibitem[Boscaini et~al.(2016)Boscaini, Masci, Rodol{\`a}, and
  Bronstein]{boscainiLearningShapeCorrespondence2016}
Boscaini, D., Masci, J., Rodol{\`a}, E., and Bronstein, M.~M.
\newblock Learning shape correspondence with anisotropic convolutional neural
  networks.
\newblock In \emph{NIPS}, 2016.

\bibitem[Bronstein et~al.(2017)Bronstein, Bruna, LeCun, Szlam, and
  Vandergheynst]{bronsteinGeometricDeepLearning2016}
Bronstein, M.~M., Bruna, J., LeCun, Y., Szlam, A., and Vandergheynst, P.
\newblock Geometric deep learning: Going beyond {{Euclidean}} data.
\newblock \emph{IEEE Signal Processing Magazine}, 2017.

\bibitem[Bruna et~al.(2014)Bruna, Zaremba, Szlam, and
  LeCun]{brunaSpectralNetworksDeep}
Bruna, J., Zaremba, W., Szlam, A., and LeCun, Y.
\newblock Spectral {{Networks}} and {{Deep Locally Connected Networks}} on
  {{Graphs}}.
\newblock In \emph{ICLR}, 2014.

\bibitem[Cohen et~al.(2018{\natexlab{a}})Cohen, Geiger, and
  Weiler]{cohenGeneralTheoryEquivariant2018}
Cohen, T., Geiger, M., and Weiler, M.
\newblock A {{General Theory}} of {{Equivariant CNNs}} on {{Homogeneous
  Spaces}}.
\newblock 2018{\natexlab{a}}.

\bibitem[Cohen \& Welling(2016)Cohen and
  Welling]{cohenGroupEquivariantConvolutional2016}
Cohen, T.~S. and Welling, M.
\newblock Group equivariant convolutional networks.
\newblock In \emph{ICML}, 2016.

\bibitem[Cohen \& Welling(2017)Cohen and Welling]{cohenSteerableCNNs2017}
Cohen, T.~S. and Welling, M.
\newblock Steerable {{CNNs}}.
\newblock In \emph{ICLR}, 2017.

\bibitem[Cohen et~al.(2018{\natexlab{b}})Cohen, Geiger, Koehler, and
  Welling]{cohenSphericalCNNs2018}
Cohen, T.~S., Geiger, M., Koehler, J., and Welling, M.
\newblock Spherical {{CNNs}}.
\newblock In \emph{ICLR}, 2018{\natexlab{b}}.

\bibitem[Cohen et~al.(2018{\natexlab{c}})Cohen, Geiger, and
  Weiler]{cohenIntertwinersInducedRepresentations2018}
Cohen, T.~S., Geiger, M., and Weiler, M.
\newblock Intertwiners between {{Induced Representations}} (with
  {{Applications}} to the {{Theory}} of {{Equivariant Neural Networks}}).
\newblock 2018{\natexlab{c}}.

\bibitem[Crane(2014)]{craneDiscreteDifferentialGeometry2014}
Crane, K.
\newblock Discrete {{Differential Geometry}}: {{An Applied Introduction}}.
\newblock 30664:\penalty0 1--6, 2014.

\bibitem[Esteves et~al.(2018)Esteves, Allen-Blanchette, Makadia, and
  Daniilidis]{estevesLearningEquivariantRepresentations2018}
Esteves, C., Allen-Blanchette, C., Makadia, A., and Daniilidis, K.
\newblock Learning {SO}(3) equivariant representations with spherical cnns.
\newblock In \emph{ECCV}, 2018.

\bibitem[Hartford et~al.(2018)Hartford, Graham, Leyton-Brown, and
  Ravanbakhsh]{hartfordDeepModelsInteractions2018}
Hartford, J.~S., Graham, D.~R., Leyton-Brown, K., and Ravanbakhsh, S.
\newblock Deep models of interactions across sets.
\newblock In \emph{ICML}, 2018.

\bibitem[He et~al.(2016)He, Zhang, Ren, and Sun]{heDeepResidualLearning2016}
He, K., Zhang, X., Ren, S., and Sun, J.
\newblock Deep {{Residual Learning}} for {{Image Recognition}}.
\newblock In \emph{CVPR}, 2016.

\bibitem[Hoogeboom et~al.(2018)Hoogeboom, Peters, Cohen, and
  Welling]{hoogeboomHexaConv2018}
Hoogeboom, E., Peters, J. W.~T., Cohen, T.~S., and Welling, M.
\newblock {{HexaConv}}.
\newblock In \emph{ICLR}, 2018.

\bibitem[Husem\"oller(1994)]{husemollerFibreBundles1994a}
Husem\"oller, D.
\newblock \emph{Fibre Bundles}.
\newblock Number~20 in Graduate Texts in Mathematics. {Springer-Verlag}, New
  York, 3rd ed edition, 1994.
\newblock ISBN 978-0-387-94087-8.

\bibitem[Hy et~al.(2018)Hy, Trivedi, Pan, Anderson, and
  Kondor]{hyPredictingMolecularProperties2018}
Hy, T.~S., Trivedi, S., Pan, H., Anderson, B.~M., and Kondor, R.
\newblock Predicting molecular properties with covariant compositional
  networks.
\newblock \emph{The Journal of Chemical Physics}, 148\penalty0 (24), 2018.

\bibitem[Ioffe \& Szegedy(2015)Ioffe and
  Szegedy]{ioffeBatchNormalizationAccelerating2015}
Ioffe, S. and Szegedy, C.
\newblock Batch normalization: Accelerating deep network training by reducing
  internal covariate shift.
\newblock \emph{arXiv:1502.03167 [cs]}, Feb 2015.
\newblock URL \url{http://arxiv.org/abs/1502.03167}.
\newblock arXiv: 1502.03167.

\bibitem[Jiang et~al.(2018)Jiang, Huang, Kashinath, Prabhat, Marcus, and
  Niessner]{jiangSphericalCNNsUnstructured2018}
Jiang, C., Huang, J., Kashinath, K., Prabhat, Marcus, P., and Niessner, M.
\newblock Spherical {{CNNs}} on {{Unstructured Grids}}.
\newblock In \emph{ICLR}, 2018.

\bibitem[Kipf \& Welling(2017)Kipf and
  Welling]{kipfSemiSupervisedClassificationGraph2017}
Kipf, T.~N. and Welling, M.
\newblock Semi-supervised classification with graph convolutional networks.
\newblock In \emph{ICLR}, 2017.

\bibitem[Kondor \& Trivedi(2018)Kondor and
  Trivedi]{kondorGeneralizationEquivarianceConvolution2018}
Kondor, R. and Trivedi, S.
\newblock On the {{Generalization}} of {{Equivariance}} and {{Convolution}} in
  {{Neural Networks}} to the {{Action}} of {{Compact Groups}}.
\newblock In \emph{ICML}, 2018.

\bibitem[Kondor et~al.(2018{\natexlab{a}})Kondor, Lin, and
  Trivedi]{kondorClebschGordanNets2018}
Kondor, R., Lin, Z., and Trivedi, S.
\newblock Clebsch\textendash{{Gordan Nets}}: A {{Fully Fourier Space Spherical
  Convolutional Neural Network}}.
\newblock In \emph{NIPS}, 2018{\natexlab{a}}.

\bibitem[Kondor et~al.(2018{\natexlab{b}})Kondor, Son, Pan, Anderson, and
  Trivedi]{kondorCovariantCompositionalNetworks2018}
Kondor, R., Son, H.~T., Pan, H., Anderson, B., and Trivedi, S.
\newblock Covariant {{Compositional Networks For Learning Graphs}}.
\newblock January 2018{\natexlab{b}}.

\bibitem[Kostelec \& Rockmore(2007)Kostelec and
  Rockmore]{kostelecSOFTFourierTransforms2007}
Kostelec, P.~J. and Rockmore, D.~N.
\newblock {{SOFT}}: {{SO}}(3) {{Fourier Transforms}}.
\newblock 2007.
\newblock URL \url{{https://www.cs.dartmouth.edu/~geelong/soft/soft20_fx.pdf}}.

\bibitem[Kunis \& Potts(2003)Kunis and Potts]{kunisFastSphericalFourier2003}
Kunis, S. and Potts, D.
\newblock Fast spherical {{Fourier}} algorithms.
\newblock 161:\penalty0 75--98, 2003.
\newblock ISSN 0377-0427.

\bibitem[Lee()]{leeIntroductionRiemannianManifolds2018}
Lee, J.
\newblock \emph{Introduction to {{Riemannian Manifolds}}}.
\newblock Graduate {{Texts}} in {{Mathematics}}. {Springer International
  Publishing}, 2 edition.
\newblock ISBN 978-3-319-91754-2.

\bibitem[Liu et~al.(2019)Liu, Yao, Choi, Ayan, and
  Karthik]{liuDeepLearning3D2019}
Liu, M., Yao, F., Choi, C., Ayan, S., and Karthik, R.
\newblock Deep {{Learning 3D Shapes}} using {{Alt}}-{{Az Anisotropic}}
  2-{{Sphere Convolution}}.
\newblock In \emph{ICLR}, 2019.

\bibitem[Lunter \& Brown(2018)Lunter and
  Brown]{lunterEquivariantBayesianConvolutional2018}
Lunter, G. and Brown, R.
\newblock An {{Equivariant Bayesian Convolutional Network}} predicts
  recombination hotspots and accurately resolves binding motifs.
\newblock \emph{Bioinformatics}, 2018.

\bibitem[Marcos et~al.(2017)Marcos, Volpi, Komodakis, and
  Tuia]{marcosRotationEquivariantVector2017}
Marcos, D., Volpi, M., Komodakis, N., and Tuia, D.
\newblock Rotation equivariant vector field networks.
\newblock In \emph{ICCV}, 2017.

\bibitem[Maron et~al.(2019)Maron, {Ben-Hamu}, Shamir, and
  Lipman]{maronInvariantEquivariantGraph2019}
Maron, H., {Ben-Hamu}, H., Shamir, N., and Lipman, Y.
\newblock Invariant and {{Equivariant Graph Networks}}.
\newblock In \emph{{{arXiv}}:1812.09902 [Cs, Stat]}, 2019.

\bibitem[Masci et~al.(2015)Masci, Boscaini, Bronstein, and
  Vandergheynst]{masciGeodesicConvolutionalNeural2015}
Masci, J., Boscaini, D., Bronstein, M.~M., and Vandergheynst, P.
\newblock Geodesic convolutional neural networks on riemannian manifolds.
\newblock \emph{ICCVW}, 2015.

\bibitem[Mudigonda et~al.(2017)Mudigonda, Kim, Mahesh, Kahou, Kashinath,
  Williams, Michalski, O'Brien, and
  Prabhat]{mudigondaSegmentingTrackingExtreme}
Mudigonda, M., Kim, S., Mahesh, A., Kahou, S., Kashinath, K., Williams, D.,
  Michalski, V., O'Brien, T., and Prabhat, M.
\newblock Segmenting and {{Tracking Extreme Climate Events}} using {{Neural
  Networks}}.
\newblock 2017.

\bibitem[Nakahara(2003)]{nakaharaGeometryTopologyPhysics2003}
Nakahara, M.
\newblock \emph{Geometry, {{Topology}}, and {{Physics}}}.
\newblock 2003.
\newblock ISBN 978-0-7503-0606-5.

\bibitem[Oyallon \& Mallat(2015)Oyallon and
  Mallat]{oyallonDeepRotoTranslationScattering2015}
Oyallon, E. and Mallat, S.
\newblock Deep roto-translation scattering for object classification.
\newblock In \emph{CVPR}, 2015.

\bibitem[Perraudin et~al.(2018)Perraudin, Defferrard, Kacprzak, and
  Sgier]{perraudinDeepSphereEfficientSpherical2018}
Perraudin, N., Defferrard, M., Kacprzak, T., and Sgier, R.
\newblock {{DeepSphere}}: {{Efficient}} spherical {{Convolutional Neural
  Network}} with {{HEALPix}} sampling for cosmological applications.
\newblock \emph{arXiv:1810.12186 [astro-ph]}, October 2018.

\bibitem[Poulenard \& Ovsjanikov(2018)Poulenard and
  Ovsjanikov]{poulenardMultidirectionalGeodesicNeural2018a}
Poulenard, A. and Ovsjanikov, M.
\newblock Multi-directional geodesic neural networks via equivariant
  convolution.
\newblock \emph{ACM Transactions on Graphics}, 2018.

\bibitem[Ravanbakhsh et~al.(2017)Ravanbakhsh, Schneider, and
  P{\'o}czos]{ravanbakhshEquivarianceParameterSharing2017}
Ravanbakhsh, S., Schneider, J., and P{\'o}czos, B.
\newblock Equivariance through parameter-sharing.
\newblock In \emph{ICML}, 2017.

\bibitem[Ronneberger et~al.(2015)Ronneberger, P.Fischer, and Brox]{RFB15a}
Ronneberger, O., P.Fischer, and Brox, T.
\newblock U-net: Convolutional networks for biomedical image segmentation.
\newblock In \emph{MICCAI}, 2015.

\bibitem[Schonsheck et~al.(2018)Schonsheck, Dong, and
  Lai]{schonsheckParallelTransportConvolution2018}
Schonsheck, S.~C., Dong, B., and Lai, R.
\newblock Parallel {{Transport Convolution}}: {{A New Tool}} for
  {{Convolutional Neural Networks}} on {{Manifolds}}.
\newblock \emph{arXiv:1805.07857 [cs, math, stat]}, May 2018.

\bibitem[Schuller(2016)]{schullerLecturesGeometricalAnatomy2016}
Schuller, F.
\newblock Lectures on the {{Geometrical Anatomy}} of {{Theoretical Physics}},
  2016.

\bibitem[Sharpe(1997)]{sharpeDifferentialGeometryCartan1997}
Sharpe, R.~W.
\newblock \emph{Differential Geometry: Cartan’s Generalization of Klein’s
  Erlangen Program}.
\newblock 1997.

\bibitem[Shoshichi~Kobayashi(1963)]{shoshichikobayashiFoundationsDifferentialGeometry1963}
Shoshichi~Kobayashi, K.~N.
\newblock \emph{Foundations of Differential Geometry (Volume 1)}.
\newblock 1963.

\bibitem[Steenrod(1951)]{steenrodTopologyFibreBundles}
Steenrod, N.
\newblock \emph{The {{Topology}} of {{Fibre Bundles}}}.
\newblock 1951.

\bibitem[Su \& Grauman(2017)Su and Grauman]{suLearningSphericalConvolution2017}
Su, Y.-C. and Grauman, K.
\newblock Flat2sphere: Learning spherical convolution for fast features from
  360 imagery.
\newblock In \emph{NIPS}, 2017.

\bibitem[Thomas et~al.(2018)Thomas, Smidt, Kearnes, Yang, Li, Kohlhoff, and
  Riley]{thomasTensorFieldNetworks2018}
Thomas, N., Smidt, T., Kearnes, S., Yang, L., Li, L., Kohlhoff, K., and Riley,
  P.
\newblock Tensor {{Field Networks}}: {{Rotation}}- and
  {{Translation}}-{{Equivariant Neural Networks}} for {{3D Point Clouds}}.
\newblock February 2018.

\bibitem[Veeling et~al.(2018)Veeling, Linmans, Winkens, Cohen, and
  Welling]{veelingRotationEquivariantCNNs2018}
Veeling, B.~S., Linmans, J., Winkens, J., Cohen, T.~S., and Welling, M.
\newblock Rotation {{Equivariant CNNs}} for {{Digital Pathology}}.
\newblock In \emph{MICCAI}, 2018.

\bibitem[Weiler et~al.(2018{\natexlab{a}})Weiler, Geiger, Welling, Boomsma, and
  Cohen]{weiler3DSteerableCNNs2018}
Weiler, M., Geiger, M., Welling, M., Boomsma, W., and Cohen, T.~S.
\newblock {3D Steerable CNNs: Learning Rotationally Equivariant Features in
  Volumetric Data}.
\newblock In \emph{NeurIPS}, 2018{\natexlab{a}}.

\bibitem[Weiler et~al.(2018{\natexlab{b}})Weiler, Hamprecht, and
  Storath]{weilerLearningSteerableFilters2018}
Weiler, M., Hamprecht, F.~A., and Storath, M.
\newblock Learning {{Steerable Filters}} for {{Rotation Equivariant CNNs}}.
\newblock In \emph{CVPR}, 2018{\natexlab{b}}.

\bibitem[Wendl(2008)]{wendlLectureNotesBundles2008}
Wendl, C.
\newblock \emph{Lecture Notes on Bundles and Connections}.
\newblock 2008.
\newblock URL
  \url{https://www.mathematik.hu-berlin.de/~wendl/connections.html}.

\bibitem[Winkels \& Cohen(2018)Winkels and Cohen]{winkels3DGCNNsPulmonary2018}
Winkels, M. and Cohen, T.~S.
\newblock {{3D G}}-{{CNNs}} for {{Pulmonary Nodule Detection}}.
\newblock In \emph{International {{Conference}} on {{Medical Imaging}} with
  {{Deep Learning}} ({{MIDL}})}, 2018.

\bibitem[Worrall \& Brostow(2018)Worrall and
  Brostow]{worrallCubeNetEquivariance3D2018}
Worrall, D.~E. and Brostow, G.~J.
\newblock Cubenet: Equivariance to 3d rotation and translation.
\newblock In \emph{ECCV}, 2018.

\bibitem[Worrall et~al.(2017)Worrall, Garbin, Turmukhambetov, and
  Brostow]{worrallHarmonicNetworksDeep2017}
Worrall, D.~E., Garbin, S.~J., Turmukhambetov, D., and Brostow, G.~J.
\newblock Harmonic {{Networks}}: {{Deep Translation}} and {{Rotation
  Equivariance}}.
\newblock In \emph{CVPR}, 2017.

\bibitem[Zaheer et~al.(2017)Zaheer, Kottur, Ravanbakhsh, Poczos, Salakhutdinov,
  and Smola]{zaheerDeepSets2017}
Zaheer, M., Kottur, S., Ravanbakhsh, S., Poczos, B., Salakhutdinov, R.~R., and
  Smola, A.~J.
\newblock Deep sets.
\newblock In Guyon, I., Luxburg, U.~V., Bengio, S., Wallach, H., Fergus, R.,
  Vishwanathan, S., and Garnett, R. (eds.), \emph{NIPS}, 2017.

\end{thebibliography}
